\documentclass[a4paper, 10 pt, conference]{ieeeconf}
\makeatletter
\let\NAT@parse\undefined
\makeatother
\IEEEoverridecommandlockouts  % This command is only needed if you want to use the \thanks command
\usepackage[pdftex]{graphicx}
\usepackage{subfig}
\usepackage{tikz}
\usetikzlibrary{shapes,backgrounds}
\usepackage{psfrag}
\usepackage{array}
\usepackage{multirow}
\usepackage{url}
\usepackage{bm}
\usepackage{bibentry}
\usepackage{gensymb} %for \degree
\usepackage{amsmath,amssymb,amsfonts,bm}
\usepackage{mathtools}
\usepackage{booktabs}
\usepackage[bottom]{footmisc}   % places footnotes at page bottom
\usepackage{etoolbox} % for \robustify and patch
\robustify{\subref}
\usepackage{algorithmic}
\usepackage{textcomp}
\def\BibTeX{{\rm B\kern-.05em{\sc i\kern-.025em b}\kern-.08em
    T\kern-.1667em\lower.7ex\hbox{E}\kern-.125emX}}
\usepackage{fontawesome}
\hypersetup{
	bookmarks=true,
	unicode=false,
	pdftoolbar=true,
	pdfmenubar=true,
	pdffitwindow=false,
	pdfstartview={FitH},
	pdftitle={Characterization of Multiple 3D LiDARs for Localization and Mapping using NDT},
	pdfauthor={A. Carballo \emph{et~al.}}
}
\usepackage{etoolbox} % for \robustify and patch
\robustify{\subref}
\usepackage{algorithmic}
\usepackage{textcomp}
\def\BibTeX{{\rm B\kern-.05em{\sc i\kern-.025em b}\kern-.08em
    T\kern-.1667em\lower.7ex\hbox{E}\kern-.125emX}}
\usepackage{fontawesome}
\usepackage[edges]{forest}

\newsavebox{\arrangebox}

\newdimen\midrulewidth
\DeclareRobustCommand*{\IEEEauthorrefmark}[1]{\raisebox{0pt}[0pt][0pt]{\textsuperscript{\footnotesize\ensuremath{\ifcase#1\or *\or \dagger\or \ddagger\or%
				\mathsection\or \mathparagraph\or \|\or **\or \dagger\dagger%
				\or \ddagger\ddagger \else\textsuperscript{\expandafter\romannumeral#1}\fi}}}}

\graphicspath{{figs-pdf/}}
\newcommand{\etal}[1][{~et~al.}]{\emph{#1}}

\newcommand{\datasetname}{LIBRE}
\newcommand{\numlidars}{12}
\newcommand{\isep}{\hspace{-3pt}\mathrel{{.}{.}\hspace{-3pt}}\nobreak}

\definecolor{folderbg}{RGB}{124,166,198}
\definecolor{folderborder}{RGB}{110,144,169}

\def\Size{4pt}
\tikzset{
	folder/.pic={
		\filldraw[draw=folderborder,top color=folderbg!50,bottom color=folderbg]
		(-1.05*\Size,0.2\Size+5pt) rectangle ++(.75*\Size,-0.2\Size-5pt);  
		\filldraw[draw=folderborder,top color=folderbg!50,bottom color=folderbg]
		(-1.15*\Size,-\Size) rectangle (1.15*\Size,\Size);
	}
}

\forestset{is file/.style={edge path'/.expanded={%
			([xshift=\forestregister{folder indent}]!u.parent anchor) |- (.child anchor)}, inner ysep=0.05pt},
	this folder size/.style={edge path'/.expanded={%
			([xshift=\forestregister{folder indent}]!u.parent anchor) |- (.child anchor) pic[solid]{folder=#1}}, inner ysep=0.01*#1},
	folder tree indent/.style={before computing xy={l=1*#1}},
	folder icons/.style={folder, this folder size=#1, folder tree indent=1.3*#1},
	folder icons/.default={12pt},
}

\begin{document}
\title{Characterization of Multiple 3D LiDARs for Localization and Mapping using Normal Distributions Transform}
\author{Alexander Carballo$^{1,4}$ %\IEEEauthorrefmark{1}\IEEEauthorrefmark{4}
\and
Abraham Monrroy$^{2,4}$ %\IEEEauthorrefmark{2}\IEEEauthorrefmark{4}
\and
David Wong$^{1,4}$ %\IEEEauthorrefmark{2}\IEEEauthorrefmark{4}
\and
Patiphon Narksri$^{2,4}$ %\IEEEauthorrefmark{1}\IEEEauthorrefmark{4}
\and
Jacob Lambert$^{2,4}$ %\IEEEauthorrefmark{2}\IEEEauthorrefmark{4}\\
\and
Yuki Kitsukawa$^{2,4}$ %\IEEEauthorrefmark{1}\IEEEauthorrefmark{4}
\and
Eijiro Takeuchi$^{2,4}$ %\IEEEauthorrefmark{1}\IEEEauthorrefmark{4}
\and
Shinpei Kato$^{3,4,5}$ %\IEEEauthorrefmark{3}\IEEEauthorrefmark{4}\IEEEauthorrefmark{5}
\and
Kazuya Takeda$^{1,2,4}$ %\IEEEauthorrefmark{1}\IEEEauthorrefmark{2}\IEEEauthorrefmark{4}% <-this % stops a space
\thanks{$^1$Institute of Innovation for Future Society, Nagoya University, Furo-cho, Chikusa-ku, Nagoya 464-8601, Japan.}%
\thanks{$^2$Graduate School of Informatics, Nagoya University, Furo-cho, Chikusa-ku, Nagoya 464-8603, Japan.}%
\thanks{$^3$Graduate School of Information Science and Technology, University of Tokyo, 7-3-1 Hongo, Bunkyo-ku, Tokyo, 113-0033, Japan.}%
\thanks{$^4$TierIV Inc., Nagoya University Open Innovation Center, 1-3, Mei-eki 1-chome, Nakamura-Ward, Nagoya, 450-6610, Japan.}%
\thanks{$^5$The Autoware Foundation, 3-22-5, Hongo, Bunkyo-ku, Tokyo, 113-0033, Japan.}%
\thanks{Email:\tt\scriptsize\underline{alexander@g.sp.m.is.nagoya-u.ac.jp}, abraham.monrroy@tier4.jp, david.wong@tier4.jp, narksri.patiphon@g.sp.m.is.nagoya-u.ac.jp, jacob.lambert@g.sp.m.is.nagoya-u.ac.jp, yuki.kitsukawa@tier4.jp,  takeuchi@g.sp.m.is.nagoya-u.ac.jp, shinpei@is.s.u-tokyo.ac.jp, kazuya.takeda@nagoya-u.jp}}% <-this % stops a space

\maketitle

\begin{abstract}
In this work, we present a detailed comparison of ten different 3D LiDAR sensors, covering a range of manufacturers, models, and laser configurations, for the tasks of mapping and vehicle localization, using as common reference the Normal Distributions Transform (NDT) algorithm implemented in the self-driving open source platform Autoware. LiDAR data used in this study is a subset of our LiDAR Benchmarking and Reference ({\datasetname}) dataset, captured independently from each sensor, from a vehicle driven on public urban roads multiple times, at different times of the day. In this study, we analyze the performance and characteristics of each LiDAR for the tasks of (1) 3D mapping including an assessment map quality based on mean map entropy, and (2) 6-DOF localization using a ground truth reference map.
\end{abstract}

\begin{keywords}
3D LiDAR, localization, mapping, normal distributions transform, LIBRE dataset
\end{keywords}

% \fi
%\IEEEpeerreviewmaketitle

\section{Introduction}
\label{s:intro}
\begin{figure}[!htb]
	\begin{center}
	\hspace*{0.65cm}\includegraphics[width=0.485\textwidth]{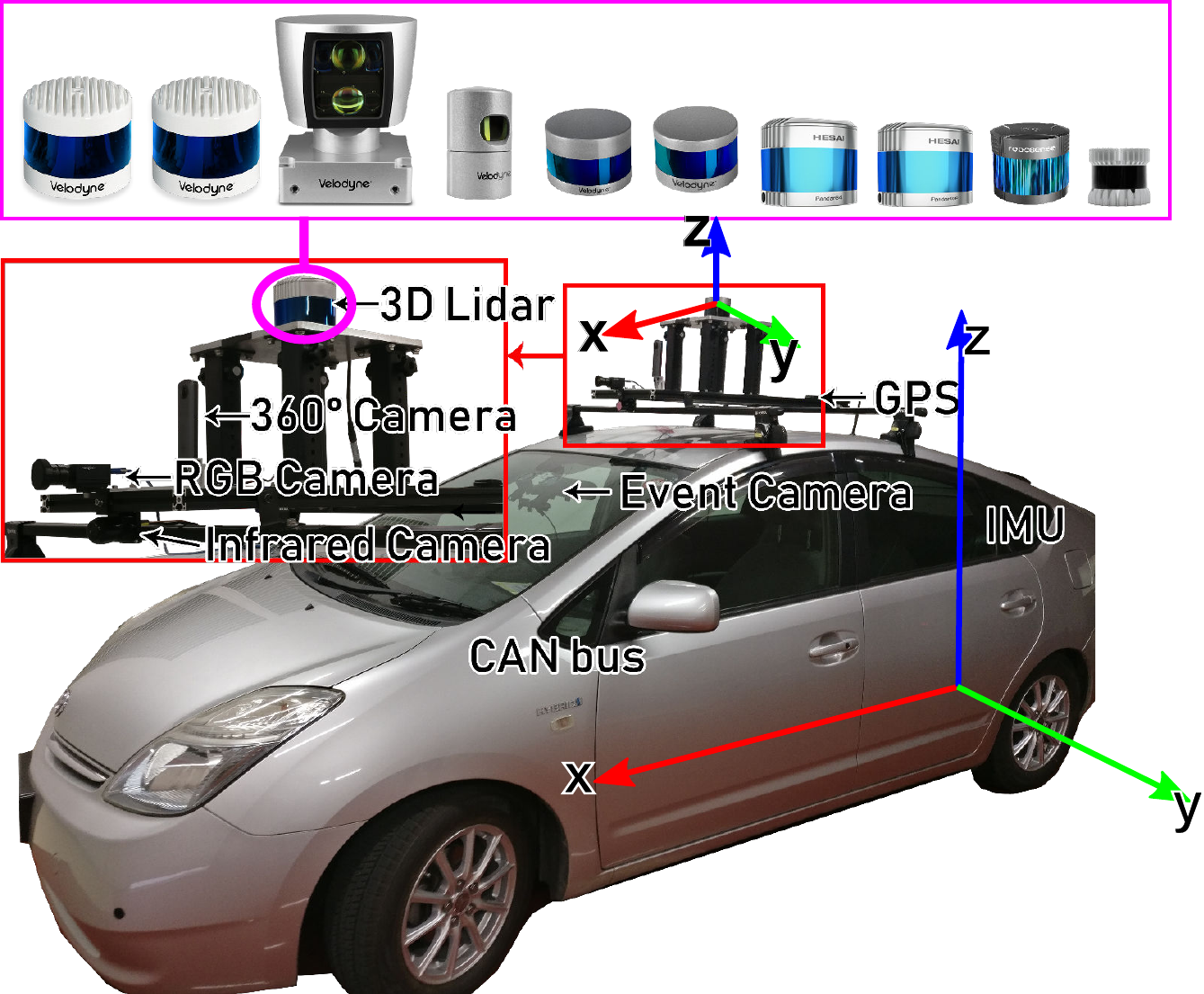}
	\caption{Instrumented vehicle used to capture dynamic traffic data in the {\datasetname} dataset and sensors used in this study.}
	\label{F:ginpuri}
	\end{center}
	\vspace{-1em}
\end{figure}
{
	\begin{table*}[t]
		\begin{center}
			\footnotesize
			\setlength{\tabcolsep}{5pt}
			\renewcommand{\arraystretch}{1.2}
			\begin{tabular}{p{1.6cm}|p{1.2cm}p{1.2cm}p{1.3cm}p{1.1cm}p{1.1cm}p{1.1cm}|p{1.3cm}p{1.3cm}|p{1.3cm}|p{1.1cm}}
				\hline\hline
				& \multicolumn{6}{c|}{Velodyne} & \multicolumn{2}{c|}{Hesai} & Ouster & RoboSense \\ %& \multicolumn{2}{c}{LeiShen}\\
				&
				\begin{minipage}{1.0cm}
					\centering
					\includegraphics[width=0.963cm]{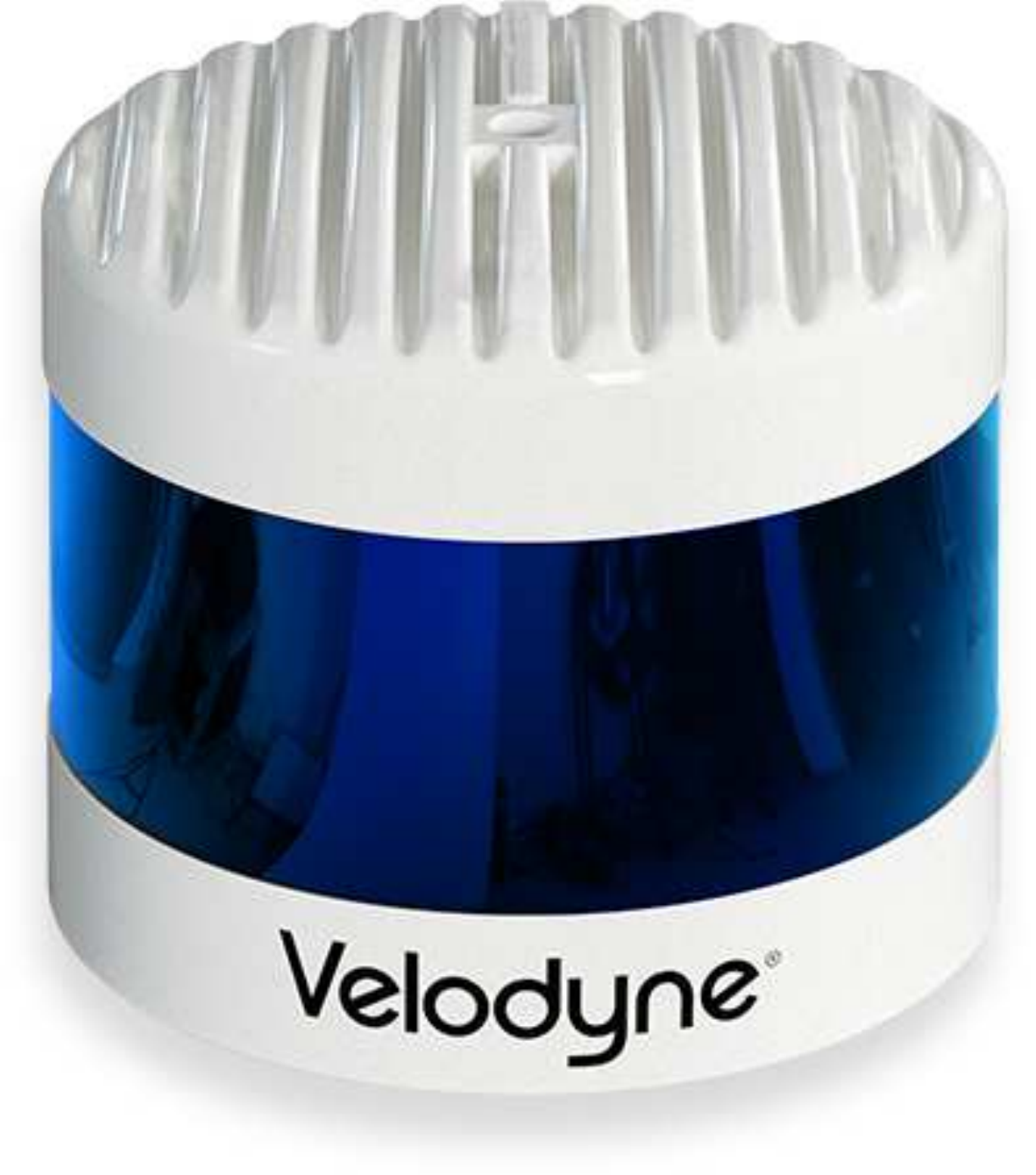}\\
					VLS-128AP$^{\bm{*}}$\cite{alphaprime}
				\end{minipage} & 
				\begin{minipage}{1.0cm}
					\centering
					\includegraphics[width=0.963cm]{vls128.pdf}\\
					VLS-128$^{\bm{*}}$\cite{vls128}
				\end{minipage} & 
				\begin{minipage}{1.3cm}
					\centering
					\includegraphics[width=1.3cm]{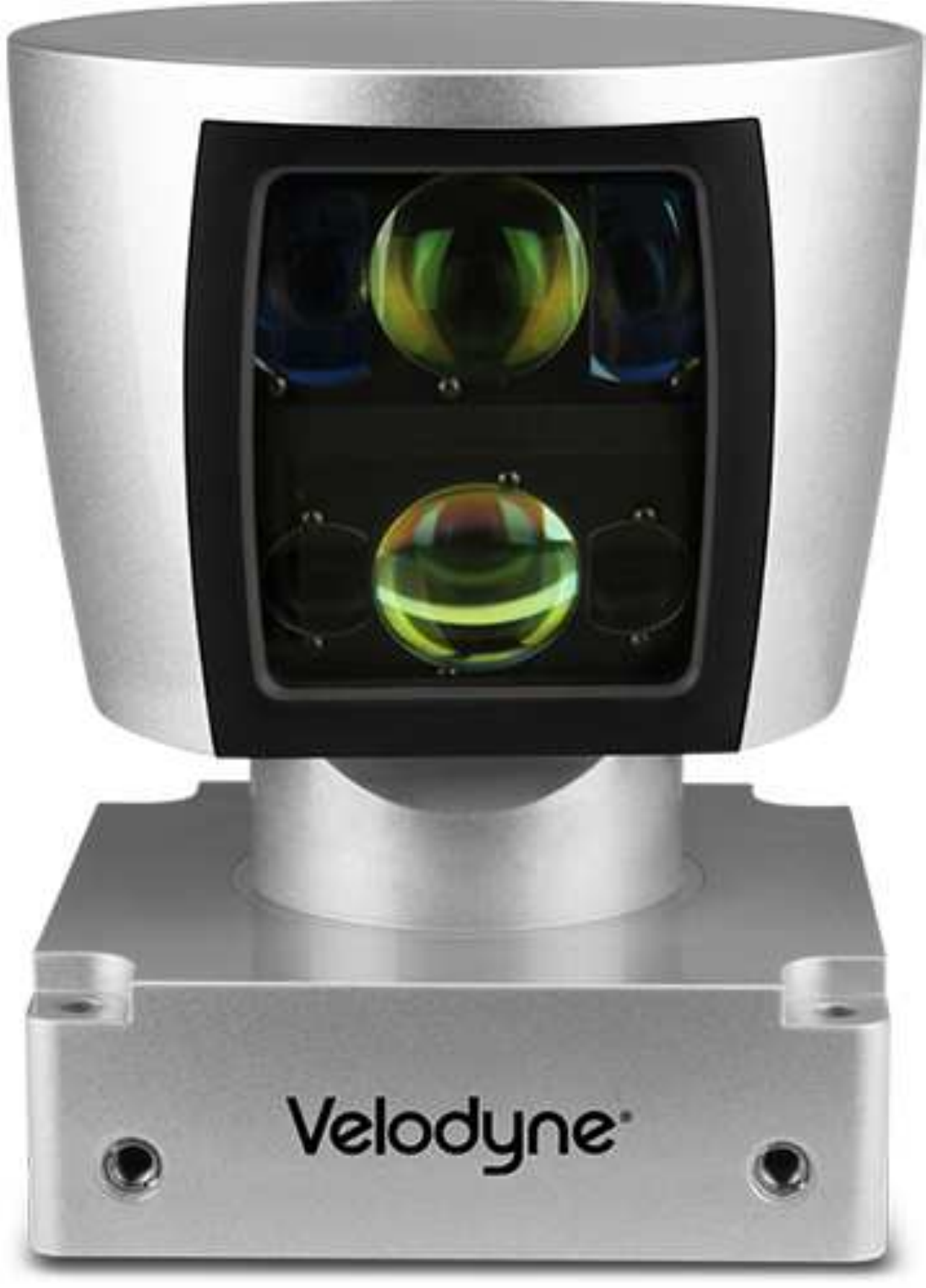}\\
					HDL-64S2\cite{hdl64s2}
				\end{minipage} &
				\begin{minipage}{0.9cm}
					\centering
					\includegraphics[width=0.4965cm]{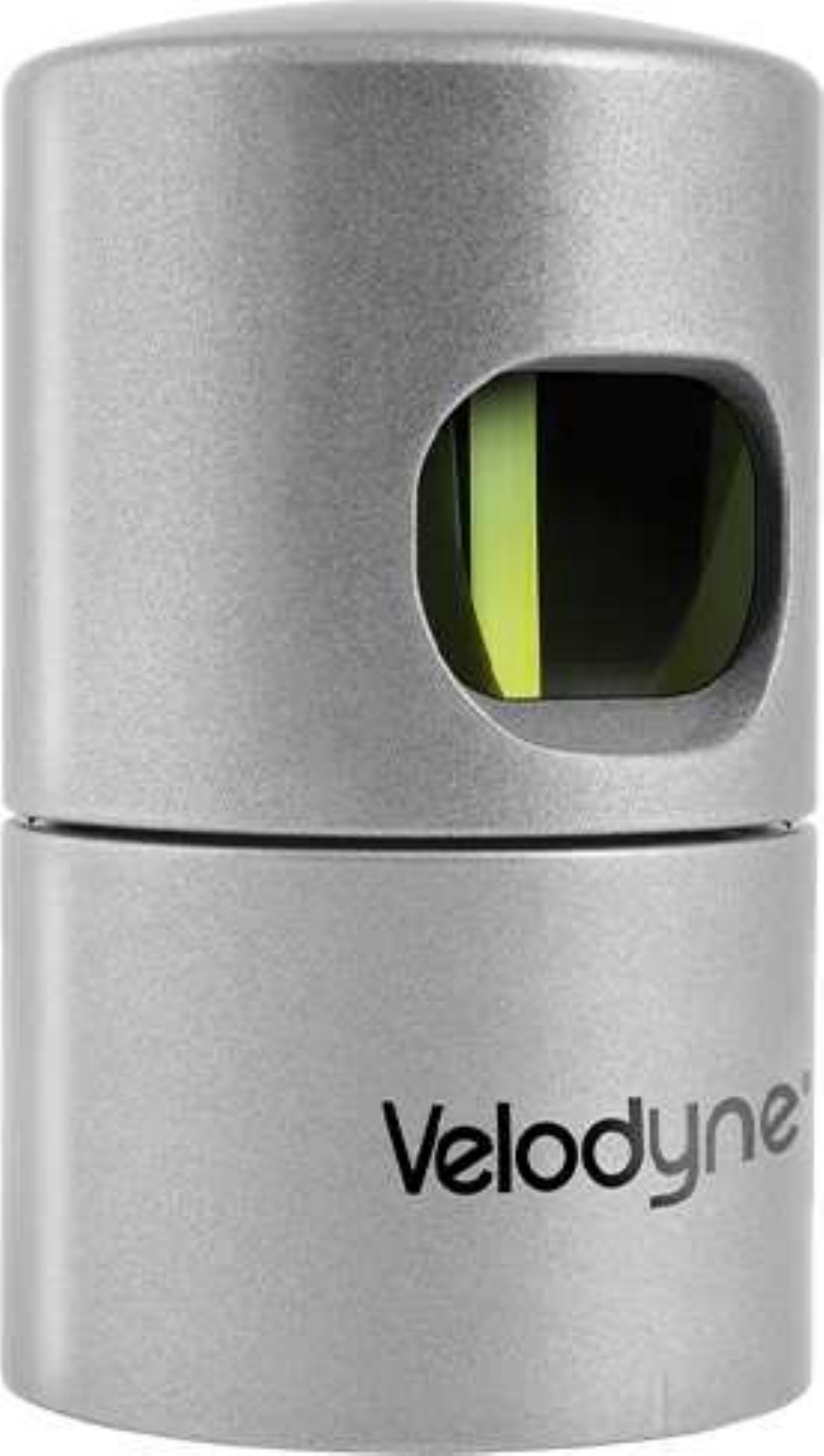}\\
					HDL-32E\cite{hdl32e}
				\end{minipage} &
				\begin{minipage}{0.7cm}
					\centering
					\includegraphics[width=0.5996cm]{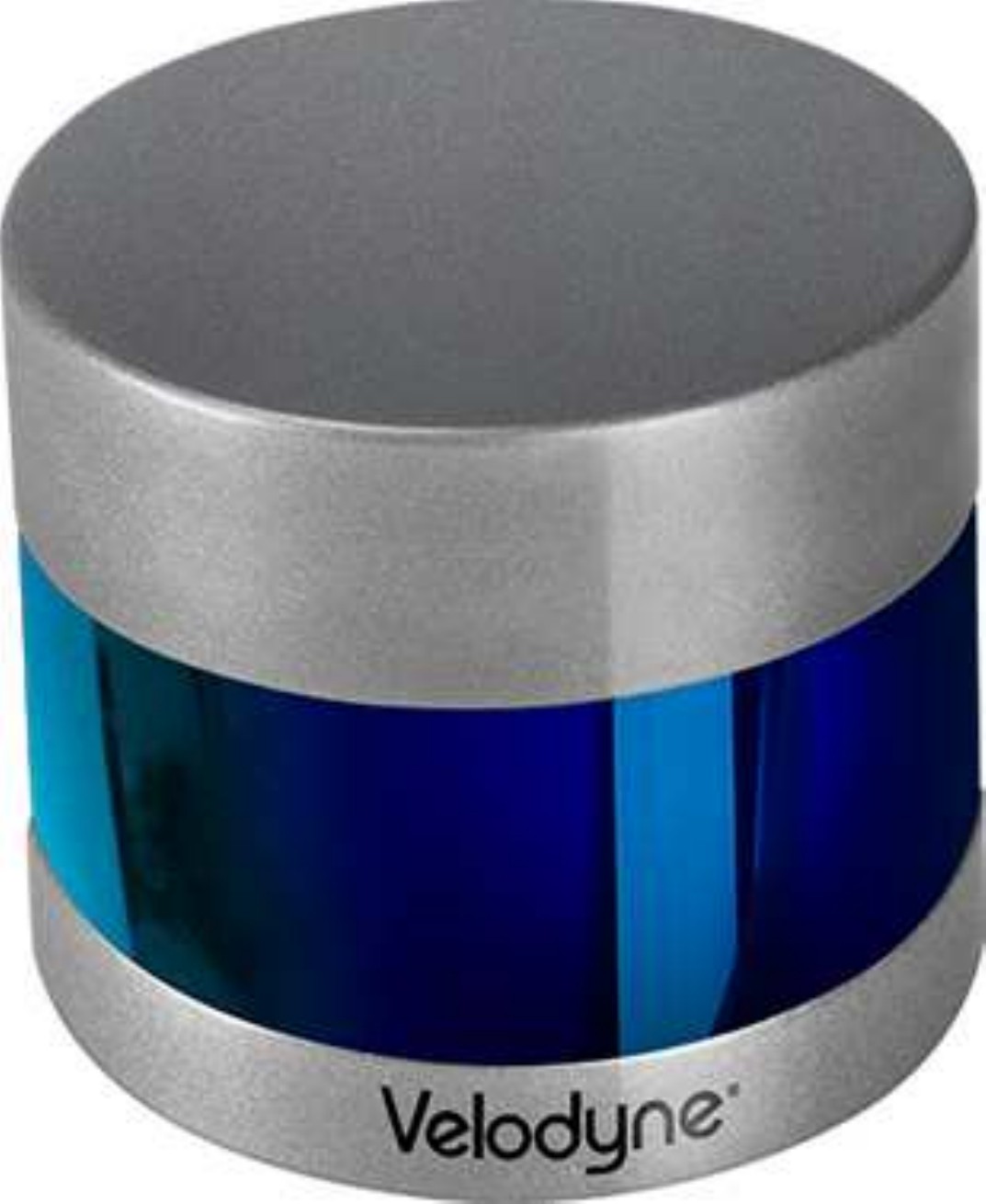}\\
					VLP-32C\cite{vlp32c}
				\end{minipage} &
				\begin{minipage}{0.7cm}
					\centering
					\includegraphics[width=0.6013cm]{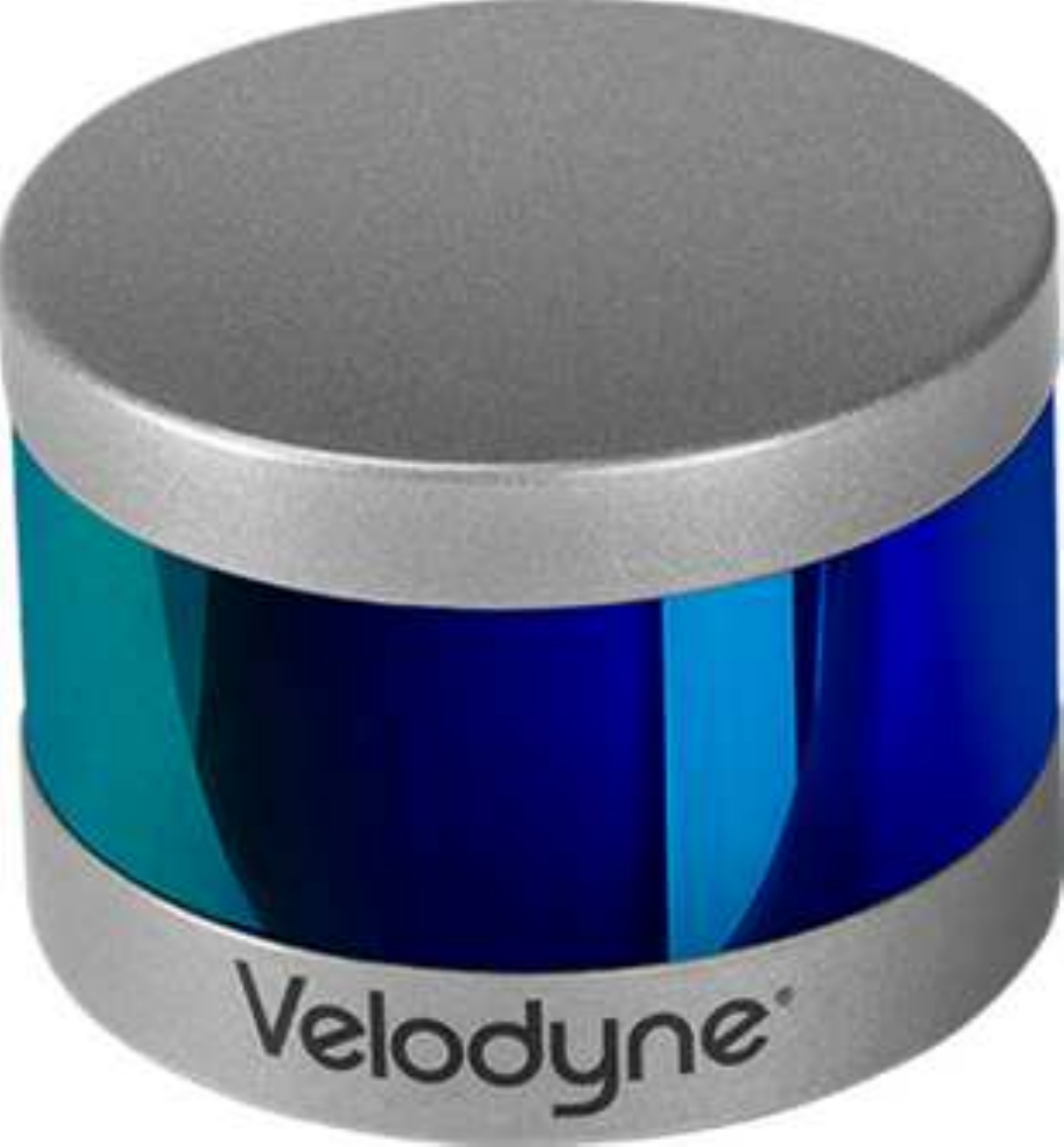}\\
					VLP-16\cite{vlp16}
				\end{minipage} &
				\begin{minipage}{0.8cm}
					\centering
					\includegraphics[width=0.6753cm]{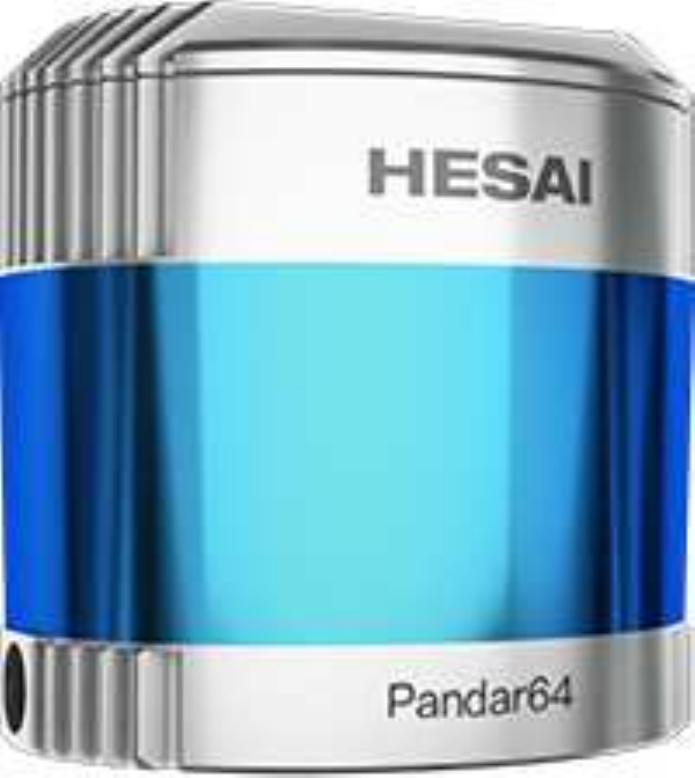}\\
					Pandar-64\cite{pandar64}
				\end{minipage} &
				\begin{minipage}{0.8cm}
					\centering
					\includegraphics[width=0.6753cm]{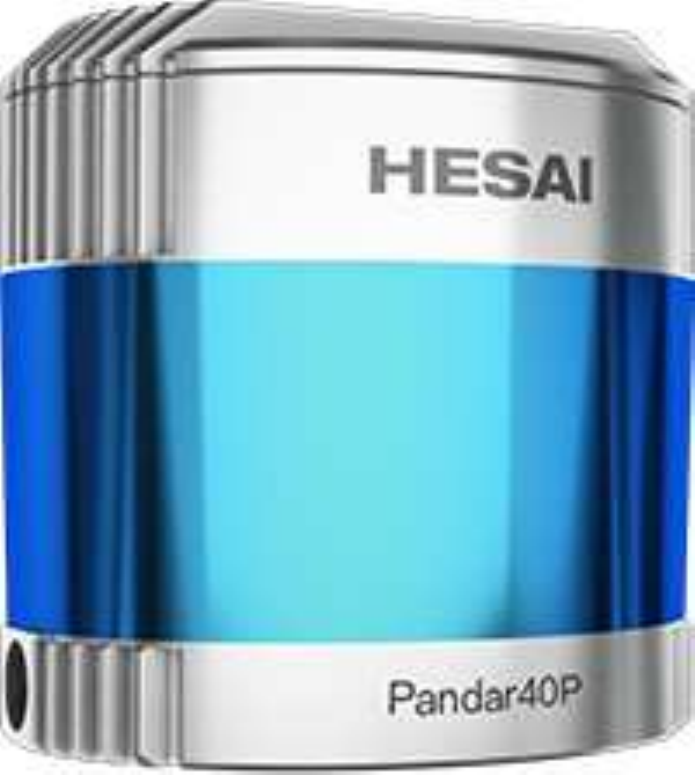}\\
					Pandar-40p\cite{pandar40p}
				\end{minipage} &
				\begin{minipage}{0.6cm}
					\centering
					\includegraphics[width=0.4948cm]{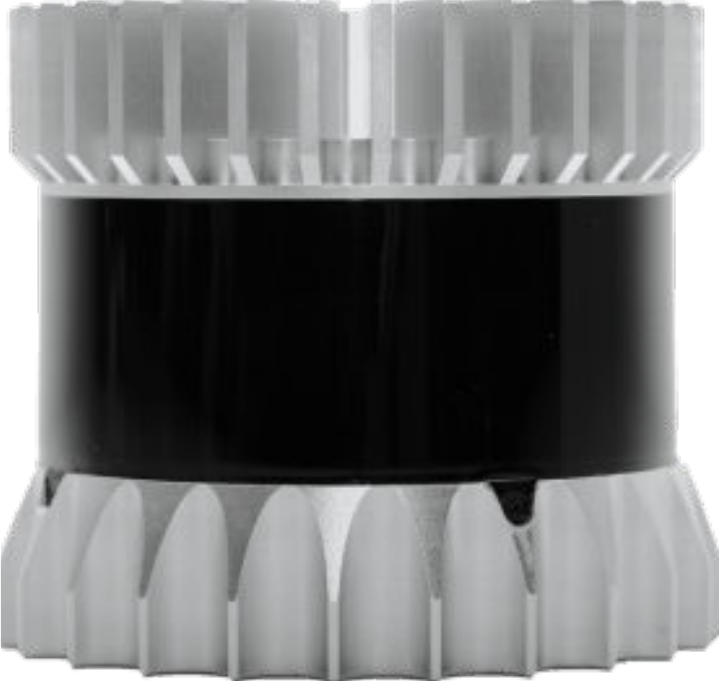}\\
					OS1-64\cite{os1}
				\end{minipage} &
				\begin{minipage}{0.7cm}
					\centering
					\includegraphics[width=0.6636cm]{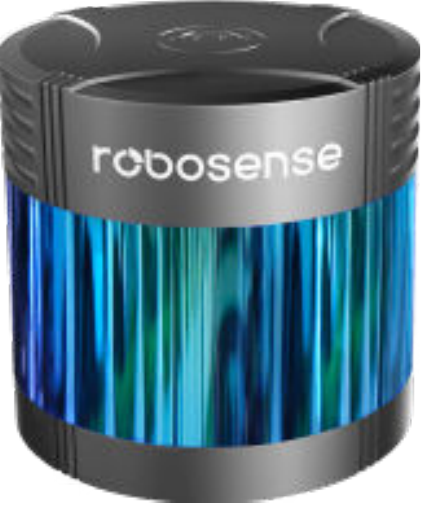}\\
					RS-Lidar32\cite{rslidar32}
				\end{minipage} \\
				\hline 
				Channels & 128 & 128 & 64 & 32 & 32 & 16 & 64 & 40 & 64 & 32 \\ 	    
				FPS[Hz] & 5-20 & 5-20 & 5-20 & 5-20 & 5-20 & 5-20 & 10,20 & 10,20 & 10,20 & 5,10,20 \\
				Precision[m] & $\pm0.03$ & $\pm0.03$ & $\pm0.02^{\bm{a}}$ & $\pm0.02$ & $\pm0.03$ & $\pm 0.03$ & $\pm0.02^{\bm{c}}$ & $\pm 0.02^{\bm{c}}$ & $\pm0.03^{\bm{d}}$ & $\pm0.03^{\bm{c}}$ \\
				Max.Range[m] & $245$ & $300$ & $120$ & $100$ & $200$ & $100$ & $200$ & $200$ & $120$ & $200$ \\
				Min.Range[m] &  &  & $3$ & $2$ & $1$ & $1$ & $0.3$ & $0.3$ & $0.8$ & $0.4$ \\ 
				vFOV[deg] & $40$ & $40$ & $26.9$ & $41.33$ & $40$ & $30$ & $40$ & $40$ & $33.2$ & $40$ \\
				vRes[deg] & ${0.11}^{\bm{b}}$ & ${0.11}^{\bm{b}}$ & ${0.33}^{\bm{a}}$ & $1.33$ & ${0.33}^{\bm{b}}$ & $2.0$ & $0.167{}^{\bm{b}}$ & $0.33{}^{\bm{b}}$ & $0.53$ & ${0.33}^{\bm{b}}$ \\
				$\lambda$[nm] & 903 & 903 & 903 & 903 & 903 & 903 & 905 & 905 & 850 & 905 \\
				$\phi$[mm] & 165.5 & 165.5 & 223.5 & 85.3 & 103 & 103.3 & 116 & 116 & 85  & 114 \\
				Weight(kg) & 3.5 & 3.5 & 13.5 & 1.0 & 0.925 & 0.830 & 1.52 & 1.52 & 0.425 & 1.17 \\
				Pointcloud fields from driver & $x$, $y$, $z$, intensity, ring, timestamp & $x$, $y$, $z$, intensity, ring, timestamp & $x$, $y$, $z$, intensity, ring & $x$, $y$, $z$, intensity, ring & $x$, $y$, $z$, intensity, ring & $x$, $y$, $z$, intensity, ring & $x$, $y$, $z$, intensity, timestamp, ring & $x$, $y$, $z$, intensity, timestamp, ring & $x$, $y$, $z$, timestamp, intensity, reflectivity, ring & $x$, $y$, $z$, intensity \\
				\hline
			\end{tabular}
			\vspace{1pt}
			\caption{LiDARs tested in this study, by manufacturer and number of channels (rings)\protect\footnotemark. Acronyms are frame rate (FPS), vertical field-of-view (vFOV), vertical resolution (vRes), laser wavelength ($\lambda$), and diameter $\phi$. 
				$\,^{\bm{a}}$Velodyne states HDL-64S2 accuracy is $\pm2cm$ for 80\% of channels, and $\pm5cm$ for the remaining; vRes for $+2\degree\isep-8.33\degree$ is $1/3\degree$ and for $-8.83\degree\isep-24.33\degree$ is $1/2\degree$. 
				$\,^{\bm{b}}$Minimum (or finest) resolution, as these sensors have variable angle difference between beams. 
				$\,^{\bm{c}}$Hesai and RoboSense state that accuracy for $0.3m\isep 0.5m$ is $\pm0.05$m, then $\pm0.02$m from $0.5m\isep 200m$. 
				$\,^{\bm{d}}$Ouster states accuracy for $0.8m\isep 2m$ is $\pm0.03m$, for $2m\isep 20m$ is $\pm0.015m$, for $20m\isep 60m$ is $\pm0.03m$, and over $60m$ is $\pm0.10m$.
				$\,^{\bm{*}}$Velodyne VLS-128 initial release (63-9480 Rev-3) and the latest Velodyne VLS-128 Alpha Prime (63-9679 Rev-1).}
			\label{tab:lidar-list}
			\vspace{-2em}
		\end{center}
	\end{table*}
	%	\footnotetext{All sensor images' not to scale and copyright owned by their respective manufacturers.}
}
State-of-the-art vehicle navigation, in particular localization and obstacle negotiation, cannot be conceived without referring to LiDARs (\emph{Light Detection And Ranging}, sometimes \emph{Light Imaging Detection And Ranging} for the image-like resolution of modern 3D sensors). Ever since the first implementations of SLAM (simultaneous localization and mapping) for robotics\cite{thrun2002robotic} and practical demonstrations on vehicles\cite{thrun2006stanley}, LRF (laser range finders) and LiDARs have had a major role to realize high accuracy 2D occupancy maps and 3D pointcloud maps. Current MMS (mobile mapping systems) employ several LiDARs together with IMUs, cameras, GNSS, odometry, to capture with the highest fidelity all the environments elements involved in driving.

The rapid developments in the field of intelligent transport systems (ITS) have created a large demand for LiDARs for obstacle negotiation, navigation, sensing in intelligent highways, and so on. Each operational design domain (ODD) defines several key requirements expected from LiDARs: measurement range, measurement accuracy, repeatability, point density, scanning speed, configurability, wavelengths, robustness to environmental changes, sensing on adverse weather, small form factors, and costs. As such, a large number of LiDAR manufacturers have emerged in recent years introducing new technologies to address different needs\cite{yole2018}. With so many different manufacturers and technologies becoming available, it is necessary to properly assess the characteristics and performance of each device. Furthermore, with LiDAR costs still remaining high, it can be difficult to select the best LiDAR in terms of cost to performance. 

We released the {\datasetname} dataset covering multiple 3D LiDARs\cite{libre,acarballo2020libre}. It features {\numlidars} LiDARs, each one a different model from diverse manufacturers, includes data from four different environments and configurations: \emph{dynamic traffic} which corresponds to traffic scenes captured from a vehicle driving on public urban roads around Nagoya University; \emph{static targets} related to objects (reflective targets, black car and mannequins) placed at known controlled distances, and measured from a fixed position; \emph{adverse weather} consisting on static objects placed at a fix location and measured from a moving vehicle while exposed to adverse conditions (fog, rain, strong light); and \emph{indirect interference} in which dynamic traffic objects are measured from a fixed position by multiple LiDARs simultaneously and exposed to indirect interference conditions. Our dataset also includes ground truth 3D pointcloud maps and vector maps (HD maps) created by a professional mobile mapping system (MMS).

Using the dynamic traffic data of our {\datasetname} dataset, this study focuses on analyzing the characteristics of ten 3D LiDARs for vehicle localization and mapping, considering the LiDAR pointcloud data alone, this is, without assistance of any other navigation and dead reckoning systems such as IMU, GNSS, odometry, or vehicle speed. To the best of our knowledge, this is the first work studying so many LiDARs for vehicle navigation, covering a range of manufacturers, models, and laser configurations (the sensors are listed in Table~\ref{tab:lidar-list}). To define a common ground for comparison, we use the Normal Distributions Transform (NDT)\cite{takeuchi2006,magnusson_ndt,magnusson2009} in the self-driving open source platform Autoware\cite{autoware-ai} (see also Kato\etal\cite{kato2018}). 

The contributions of this work are summarized as follows:
\begin{itemize}
	\item A unique study of the properties for multiple 3D LiDARs for creation of 3D maps and 6-DOF localization. 
	\item Evaluation of 3D map quality using the mean map entropy (MME) and mean plane variance (MPV) scores for each LiDAR.
%	\item Ablation studies to explore the detection range of each LiDAR versus localization results.
\end{itemize} 

This paper is structured as follows: Section~\ref{s:ndt} revisits NDT scan matching and its properties. Section~\ref{s:dyamic} describes the dynamic traffic data and experimental procedures in this study. Sections~\ref{s:mapping} and \ref{s:locoalization} discuss the actual result for mapping localization and localization. Finally, this paper is concluded in Section~\ref{s:concl}.
\footnotetext{All sensor images' not to scale and copyright owned by their respective manufacturers.}

{
	\begin{figure}[!htb]
		\centering
		\subfloat[][]{
			\includegraphics[width=0.35\textwidth]{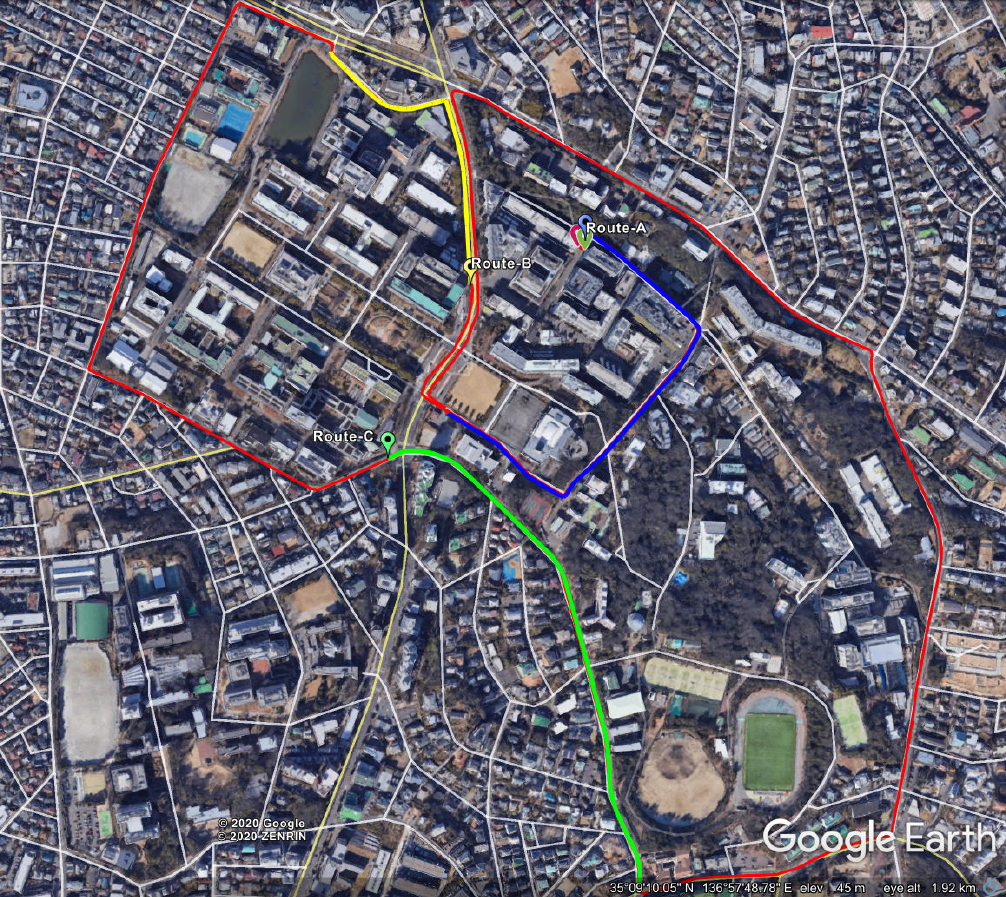}
			\label{F:libremap-a}
		}\\
		\subfloat[][]{
			\includegraphics[width=0.35\textwidth]{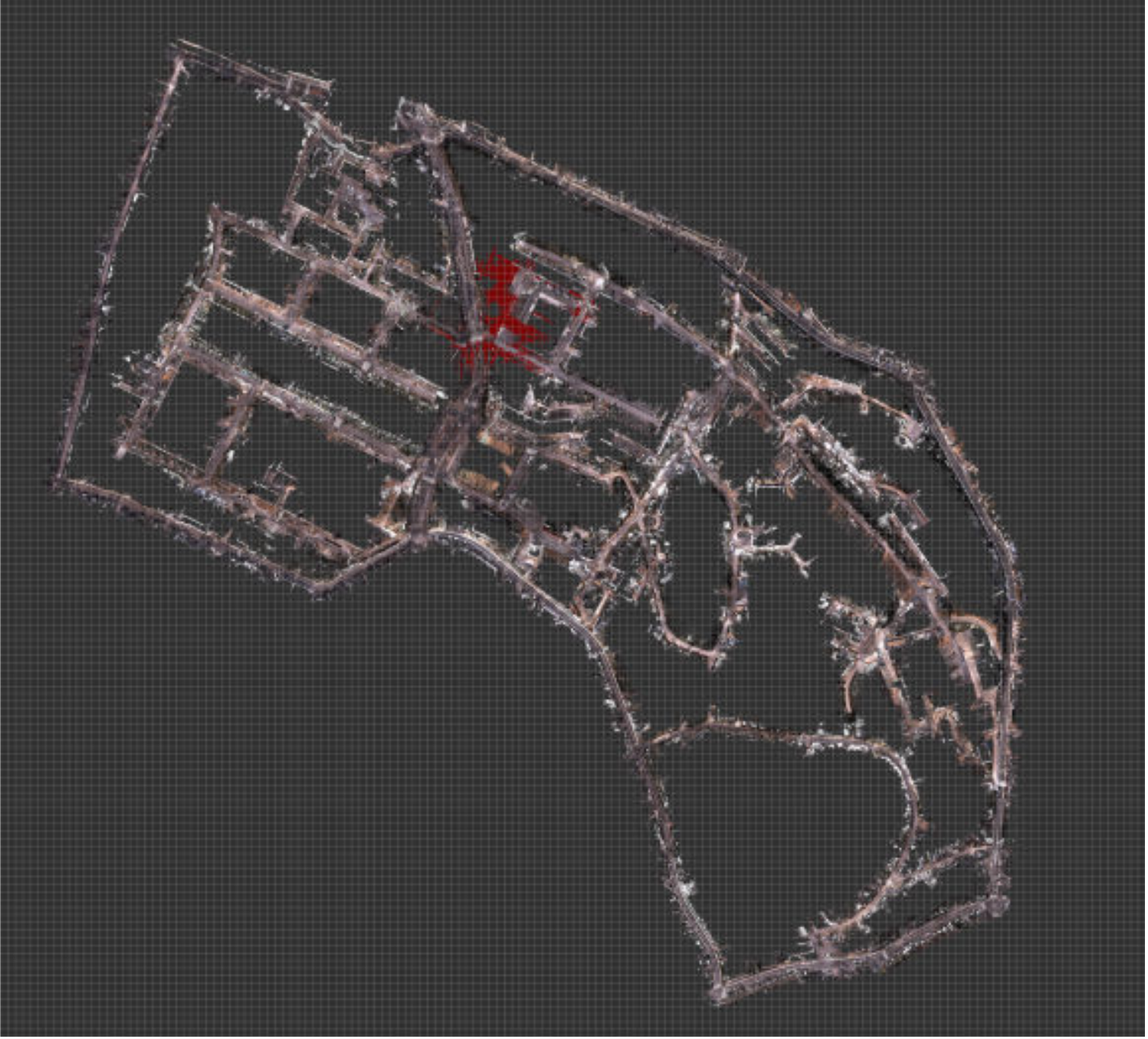}
			\label{F:libremap-b}
		}
		\caption[]%
		{Map of the dynamic environment included in the dataset: \subref{F:libremap-a} general map\protect\footnotemark, trajectory colored in red corresponds to the complete drive starting at location marked with {\color{red}{\faMapMarker}}, and ending at location marked with {\color{lime!80!blue}{\faMapMarker}}. Trajectory colored in blue with starting location marked with {\color{blue}{\faMapMarker}} corresponds to route A, yellow trajectory with start location marked with {\color{yellow}{\faMapMarker}} is route B, and trajectory colored in green with starting location marked by {\color{lime!60!green}{\faMapMarker}} is route C. \subref{F:libremap-b} is the ground truth pointcloud map (grid cell size 10\,m) created by a MMS. }
		\label{F:libremap}
		\vspace{-1em}
	\end{figure}
	\footnotetext{Google map is available at \url{https://drive.google.com/open?id=1fFja7B29xmidldVJPWLKSK0UCrXhy47N&hl=en&z=17}}
}

\section{Normal Distributions Transform}
\label{s:ndt}
\subsection{NDT Scan Matching}
\label{ss:ndtscanmatching}
The Normal Distributions Transform (NDT) scan matching method is attributed to Biber\etal\cite{biber_ndt}. In NDT, the reference scan (pointcloud) is divided into uniform 3D grid, and each cell (voxel) keeps the mean and distribution of the sub-pointcloud assigned to it, thus the ND voxels represent normal distributions.

Following the notation by Takeuchi\etal\cite{takeuchi2006}, the reference pointcloud (map or model) $\bm{M}$ is a vector of $M$ points $\bm{x}_i=\langle{x_i,y_i,z_i}\rangle^\mathsf{T}$ with $i\in[1..M]$, and $x_{k}$ is the $k$-th ND voxel with $M_{k}$ points. The $k$-th ND voxel mean $\bm{p}_k$ and covariance matrix $\bm{\Sigma}_k$ are given by Eq.~\ref{eq:mean}.

\begin{eqnarray}
\bm{p}_k&=&\frac{1}{M_k}\sum_{i=1}^{M_k}\bm{x}_{ki} \nonumber \\
\bm{\Sigma}_k&=&\frac{1}{M_k}\sum_{i=1}^{M_k} (\bm{x}_{ki}-\bm{p}_k) (\bm{x}_{ki}-\bm{p}_k)^\mathsf{T} \label{eq:mean}
\end{eqnarray}

Given an input scan (pointcloud) $\bm{X}$ with $N$ points, $\bm{x}_i$ with $i\in[1..N]$, the 3D coordinate transformation of $\bm{X}$ into $\bm{M}$ is achieved by ${\bm{x}'}_i=\bm{R}\bm{x}_i+\bm{t}'$, with $\bm{R}$ the rotation matrix and $\bm{t}'$ the translation vector. Thus, the pose transformation parameters (translation and rotation) to estimate are $\bm{t}=\langle{t_{x},t_{y},t_{z},t_{roll},t_{pitch},t_{yaw}}\rangle$. Evaluation of fitness between the transformed input cloud $\overline{\bm{X}}$ using the parameters $\bm{t}$, and the reference map $\bm{M}$, represented as ND voxels, is done with Eq.~\ref{eq:eval}.
\begin{equation}
E(\bm{X},\bm{t})=\sum_{i}^{N}\exp{\frac{-({\bm{x}'}_i-\bm{p}_i)^\mathsf{T}\bm{\Sigma}_i^{-1}({\bm{x}'}_i-\bm{p}_i)}{2}}
\label{eq:eval}
\end{equation}
A high value of $E(\bm{X},\bm{t})$ means both the input cloud and the reference map are well aligned. Newton's nonlinear function optimization is utilized to find $\bm{t}$ such that $E(\bm{X},\bm{t})$ is maximized. Therefore, we minimize the function $f(\bm{t})=-E(\bm{X},\bm{t})$. Parameters vector $\bm{t}$ is updated using Eq.~\ref{eq:update}.
\begin{equation}
\bm{t}_{new}=\bm{t}-\bm{H}^{-1}\bm{g} \label{eq:update}
\end{equation}
where $\bm{g}$ and $\bm{H}$ are the partial differential and second order partial differential of the optimizing function $f$. Details of the derivation of these values are given in \cite{takeuchi2006}, the work by Magnusson\cite{magnusson2009} includes a very derivation. Magnusson\etal\cite{magnusson_ndt} and Sobreira\etal\cite{sobreira2019map} present detailed comparisons of performance of NDT versus the Iterative Closest Point (ICP) algorithm and others.

\subsection{NDT Evaluation Metrics}
\subsubsection{Iteration}
Iteration corresponds to the number of cycles (and therefore processing time) of the Newton's iterative method until achieving matching convergence. If the initial guess for the transformation parameters $\bm{t}$ is close enough to the actual transformation, then the number of iterations is small. 

\subsubsection{Fitness Score}
Fitness Score is the degree of correspondence between two scans, obtained by the average sum of distances between closest points. When this score is small then correspondence between pointclouds is high. However, if the reference cloud lacks areas which are part of the input cloud, the distance between closest neighbors increases and so this score.

\subsubsection{Transformation Probability}
Transformation Probability, although strictly speaking not a probability, is the score of one point, obtained by dividing the fitness score by the number of points $N$ of the input scan.

\subsection{NDT Precision and Performance Factors}
\subsubsection{Input Cloud Down-sampling}
The input cloud of conventional LiDARs may consist on tens to hundreds of thousands points. As such, the time to achieve convergence will increase with the size of the input cloud and may limit real-time response. Down-sampling of the input cloud reduces points and complexity for matching.

\subsubsection{Resolution of Reference Map}
In a similar way, the complexity of the map affects NDT matching performance. Resolution corresponds to the size of the ND voxels, a large resolution improves processing time but reduces features and so matching becomes unstable. On the other hand, very small resolution means higher processing time and causes incorrect associations between input cloud and very close ND voxels, therefore accuracy is affected.

\subsubsection{VoxelGrid Filter}
Similar to the map resolution, the voxel grid filter is a down-sampling method applied on the input scan. A 3D grid is created on the input cloud, the local cloud in a voxel is replaced by the centroid (for the fields of $x$, $y$, $z$ and intensity). Voxel grid filter preserves the general coverage (in distance) of the input cloud and reduces noise, but does not preserve the ring structure: the centroid may correspond to a point in between two rings unrelated to the original scan. By reducing the input cloud complexity according to the voxel grid cell size, the number of iterations to achieve convergence may be reduced.  

\subsubsection{Number of LiDAR Beams}
As show in Table~\ref{tab:lidar-list}, there are multiple 3D LiDARs available for vehicle navigation, and one of their distinctive attributes is the number of beams (lasers or channels) and their distribution. A high number of beams and a fine horizontal resolution mean the size of the input pointcloud is large and thus more time is necessary for processing. Finer vertical angular resolutions mean better estimation of the voxel grid vertical centroid.

\subsubsection{Matching Initialization}
It is important to define the initial position and pose before a matching cycle with new scan data. As the vehicle moves continuously over time, the position computed in the previous cycle is used as a guess for the new cycle. In addition, the use of dead reckoning such as speed, acceleration, odometry and other navigation systems such as IMU and GNSS, can help improve the estimation of the initial position before the next cycle.

\section{Multiple LiDAR Dynamic Traffic Data}
\label{s:dyamic}
\subsection{Data Collection}
\label{ss:datacollection}
The target was to collect data in a variety of traffic conditions, including different type of environments, varying density of traffic and times of the day. We drove our instrumented vehicle three times per day and collected data for the following key time periods:
\begin{itemize}
	\item Morning (9am-10am)
	\begin{itemize}
		\item Pedestrian traffic: medium-low
		\item Vehicle traffic: high
		\item Conditions: people commuting to work, university students and staff arriving on the campus. Clear to overcast weather.
	\end{itemize}
	\item Noon (12pm-1pm)
	\begin{itemize}
		\item Pedestrian traffic: high
		\item Vehicle traffic: medium-low
		\item Conditions: large number of university students and staff heading to and from the various cafeterias and restaurants around Campus. These are largely accessible by walking, so vehicle traffic is relatively low around this time. Clear to overcast weather.
	\end{itemize}
	\item Afternoon (2pm-4pm)
	\begin{itemize}
		\item Pedestrian traffic: low
		\item Vehicle traffic: medium-low
		\item Conditions: foot traffic around campus is low during this busy work and class period, and vehicle density is also relatively normal or low. Clear to overcast weather.
	\end{itemize}
\end{itemize}
Fig.~\ref{F:ginpuri} shows the vehicle used for data capture. The 3D LiDAR on top was replaced for each experiment only after the three time periods were recorded, only one LiDAR was used at a time to avoid noise due to mutual interference. Data from other sensors (RGB camera, IR camera, 360$\degree$ camera, event camera, IMU, GNSS, CAN) was also recorded together with LiDAR data, together with timestamps, using ROS\cite{ros2009}. In addition, we collected calibration data for each new LiDAR setup to perform extrinsic LiDAR to camera calibration, using a checkerboard and various other points of interest. Clear lighting conditions were ensured to record such data. 

Fig.~\ref{F:libremap} shows the main trajectory followed to record dynamic traffic data. Our dataset also offers a reference pointcloud map, created by a professional mobile mapping system (MMS), and which includes 3D coordinates and RGB data. Vector map files (HD map) for public road outside of the Nagoya University campus, are also provided.

\subsection{Evaluation Routes}
\label{ss:routes}
From the main trajectory shown in Fig.~\ref{F:libremap}, we selected 3 different routes for this study:
\begin{itemize}
	\item Route A (blue): about 749\,m of private roads inside Nagoya University, narrow, surrounded by trees and buildings, gentle slopes; maximum velocity allowed inside the campus is 30\,km/h.
	\item Route B (yellow): about 475\,m of public urban roads surrounding Nagoya University campus, wide, multiple lanes, large intersections, traffic signs and marks, mostly flat; maximum velocities between 40\,km/h to 50\,km/h. 
	\item Route C (green): about 797\,m of public urban roads surrounding Nagoya University campus, two lanes, intersections, hilly (7.6\% average slope); maximum velocity 40\,km/h. 
\end{itemize}

For each route, we used the previously recorded sensor LiDAR log data included as part of our {\datasetname} dataset. To keep relative consistency in diving conditions, we chose the morning time rides for this study. In total, thirty (ten LiDARs times 3 routes) different evaluations for 3D map creation and for vehicle localization.

\begin{table*}[t]
	\begin{center}
		\begin{tabular}{l|lrrrrrrrrrr}
			\hline\hline
			& Sensor & Beams & N. points & Drive(s) & N. scans & Mean iter. & Std. iter. & Mean Fit.Sc. &  Std. Fit.Sc. & MME & MPV\\
			\hline
			\multirow{10}{*}{A} 
			& VLS128ap & 128 & 160390220 & 204 & 1958 & 5.3805 & 8.3893 & 0.5175 & 0.0430 &  & \\
			& VLS128 & 128 & 126705326 & 160 & 1586 & 6.1513 & 8.4786 & 0.5453 & 0.0335 & 0.5701 & 0.4045\\
			& Pandar64 & 64 & 134901915 & 158 & 1571 & 6.1948 & 10.1055 & 0.5154 & 0.0315 &  & \\
			& OS1-64 & 64 & 60912465 & 149 & 1486 & 4.3419 & 8.3900 & 0.5254 & 0.0304 & 0.5849 & 0.3851\\
			& HDL-64S2 & 64 & 78402509 & 177 & 1764 & 5.9195 & 8.0350 & 0.5265 & 0.0481 & \textbf{0.2976} & \textbf{0.2900}\\
			& Pandar40P & 40 & 81345147 & 153 & 1527 & 5.5200 & 8.7917 & 0.5262 & 0.0359 & 0.4923 & 0.3945\\
			& VLP-32C & 32 & 29679934 & 155 & 1528 & 3.1041 & 4.9624 & 0.5233 & 0.0379 & 0.4772 & 0.3867\\
			& RS-lidar32 & 32 & 30053527 & 153 & 1505 & 4.0339 & 6.9112 & 0.5271 & 0.0404 & 0.5283 & 0.3944\\
			& HDL-32E & 32 & 35845160 & 155 & 1547 & 3.9774 & 6.9487 & 0.5236 & 0.0423 & \textbf{0.0299} & \textbf{0.2896}\\
			& VLP-16 & 16 & 13215302 & 147 & 1447 & \textbf{2.3918} & 2.8995 & 0.5355 & 0.0406 & 0.4338 & 0.3823\\
			\hline
			\multirow{10}{*}{B} 
			& VLS128ap & 128 & 93933095 & 145 & 1388 & 3.3927 & 5.3527 & 0.5398 & 0.0331 & 0.8185 & 0.4822\\
			& VLS128 & 128 & 73962681 & 150 & 1494 & 3.2088 & 4.5320 & 0.5873 & 0.0244 & 0.6429 & 0.4347\\
			& Pandar64 & 64 & 80049525 & 289 & 2901 & 2.4430 & 2.6135 & 0.5200 & 0.0231 & 0.5289 & 0.4500\\
			& OS1-64 & 64 & 37105260 & 80 & 798 & 3.0727 & 6.0023 & 0.5328 & 0.0227 & 0.6937 & 0.4061\\
			& HDL-64S2 & 64 & 44907042 & 93 & 924 & 5.0498 & 7.7395 & 0.5187 & 0.0313 & \textbf{0.4298} & \textbf{0.3357}\\
			& Pandar40P & 40 & 46905632 & 148 & 1477 & 2.9255 & 4.6709 & 0.5224 & 0.0296 & 0.5788 & 0.4254\\
			& VLP-32C & 32 & 17055829 & 135 & 1345 & 2.4409 & 3.4703 & 0.5344 & 0.0341 & 0.5318 & 0.4150\\
			& RS-lidar32 & 32 & 18378976 & 134 & 1318 & 2.6381 & 4.4722 & 0.5486 & 0.0325 & 0.5865 & 0.4215\\
			& HDL-32E & 32 & 20332647 & 72 & 716 & 3.3254 & 5.9123 & 0.5371 & 0.0362 & \textbf{0.2752} & \textbf{0.3340}\\
			& VLP-16 & 16 & 7550326 & 128 & 1263 & \textbf{2.2486} & 2.8580 & 0.5372 & 0.0352 & 0.5446 & 0.4097\\
			\hline
			\multirow{10}{*}{C} 
			& VLS128ap & 128 & 172433639 & 80 & 767 & 7.7405 & 13.0467 & 0.5626 & 0.0496 &  & \\
			& VLS128 & 128 & 118462134 & 90 & 895 & 6.0335 & 9.0134 & 0.6249 & 0.0447 & 0.5407 & 0.4506\\
			& Pandar64 & 64 & 139121463 & 88 & 877 & 5.2201 & 8.8084 & 0.5391 & 0.0534 &  & \\
			& OS1-64 & 64 & 56799362 & 143 & 1427 & 2.5200 & 4.2354 & 0.5473 & 0.0341 & 0.6532 & 0.4188\\
			& HDL-64S2 & 64 & 68186933 & 173 & 1723 & 3.8938 & 5.8987 & 0.5114 & 0.0495 & 0.4340 & \textbf{0.3401}\\
			& Pandar40P & 40 & 71191899 & 96 & 957 & 4.3386 & 8.4673 & 0.5490 & 0.0535 & 0.5515 & 0.4615\\
			& VLP-32C & 32 & 29788546 & 87 & 866 & 3.1166 & 5.7595 & 0.5435 & 0.0688 & 0.4193 & 0.4281\\
			& RS-lidar32 & 32 & 27233475 & 154 & 1515 & 2.5300 & 3.3160 & 0.5346 & 0.0564 & 0.4301 & 0.4106\\
			& HDL-32E & 32 & 28122526 & 94 & 936 & 3.0962 & 5.6704 & 0.5566 & 0.0482 & \textbf{0.2365} & 0.3502\\
			& VLP-16 & 16 & 15153288 & 257 & 2524 & \textbf{2.0959} & 0.9073 & 0.5216 & 0.0382 & 0.5298 & 0.4145\\
			\hline
		\end{tabular}
		\caption{Mapping results for each route and each LiDAR.}
		\label{tab:mapping}
		\vspace{-2em}
	\end{center}
\end{table*}
\begin{figure}[!htb]
	\centering
	\subfloat[][Route A]{
		\includegraphics[width=0.35\textwidth]{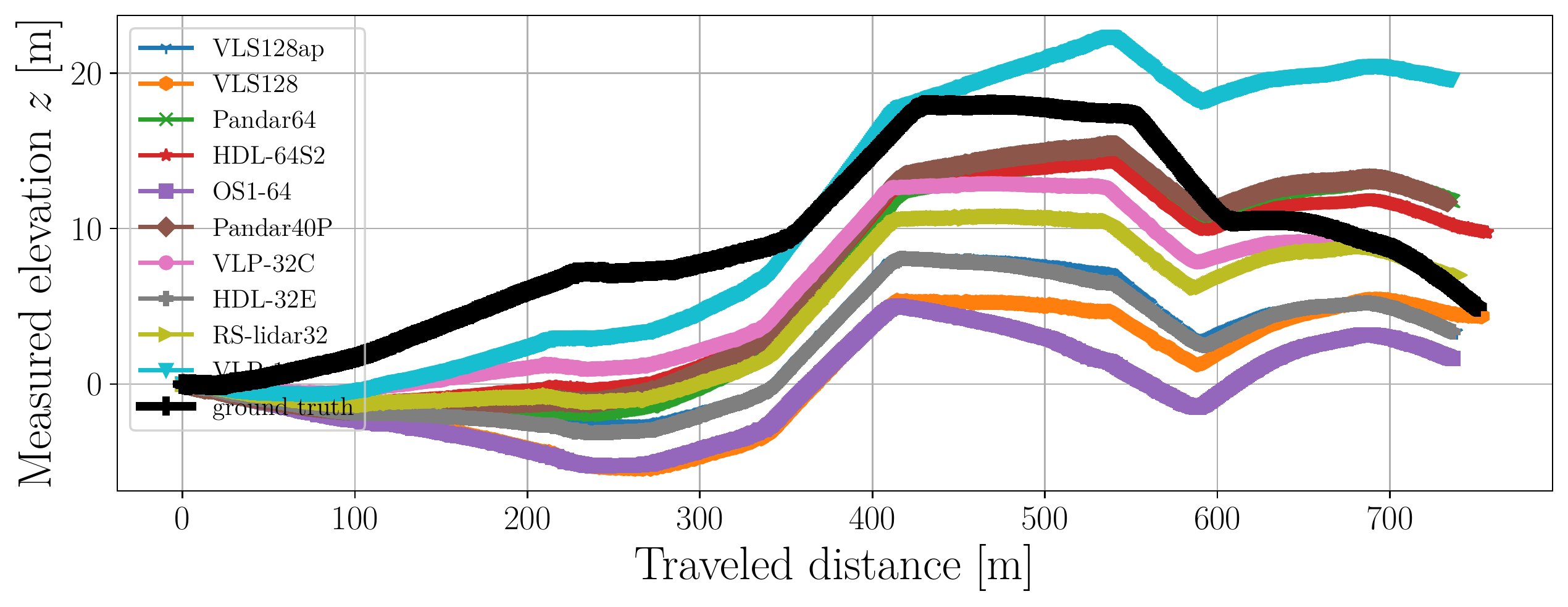}
		\label{F:elevation-a}
	}\\
	\subfloat[][Route B]{
		\includegraphics[width=0.35\textwidth]{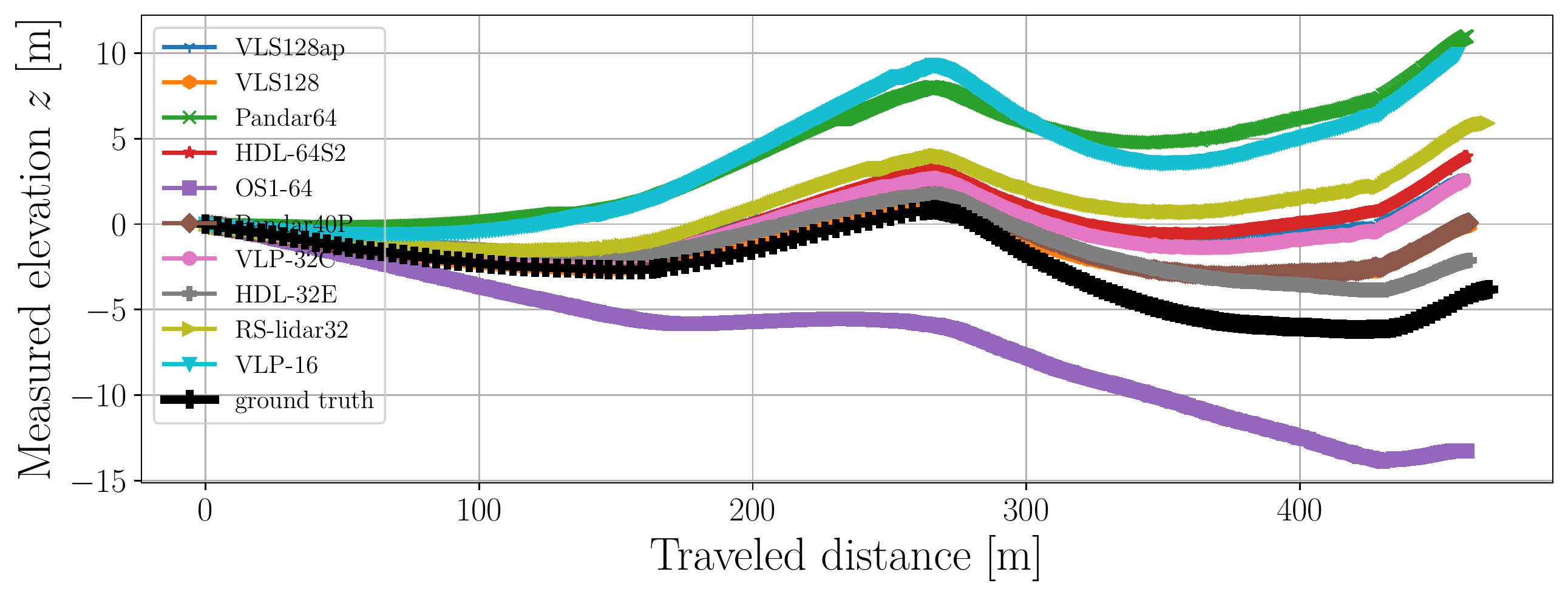}
		\label{F:elevation-b}
	}\\
	\subfloat[][Route C]{
		\includegraphics[width=0.35\textwidth]{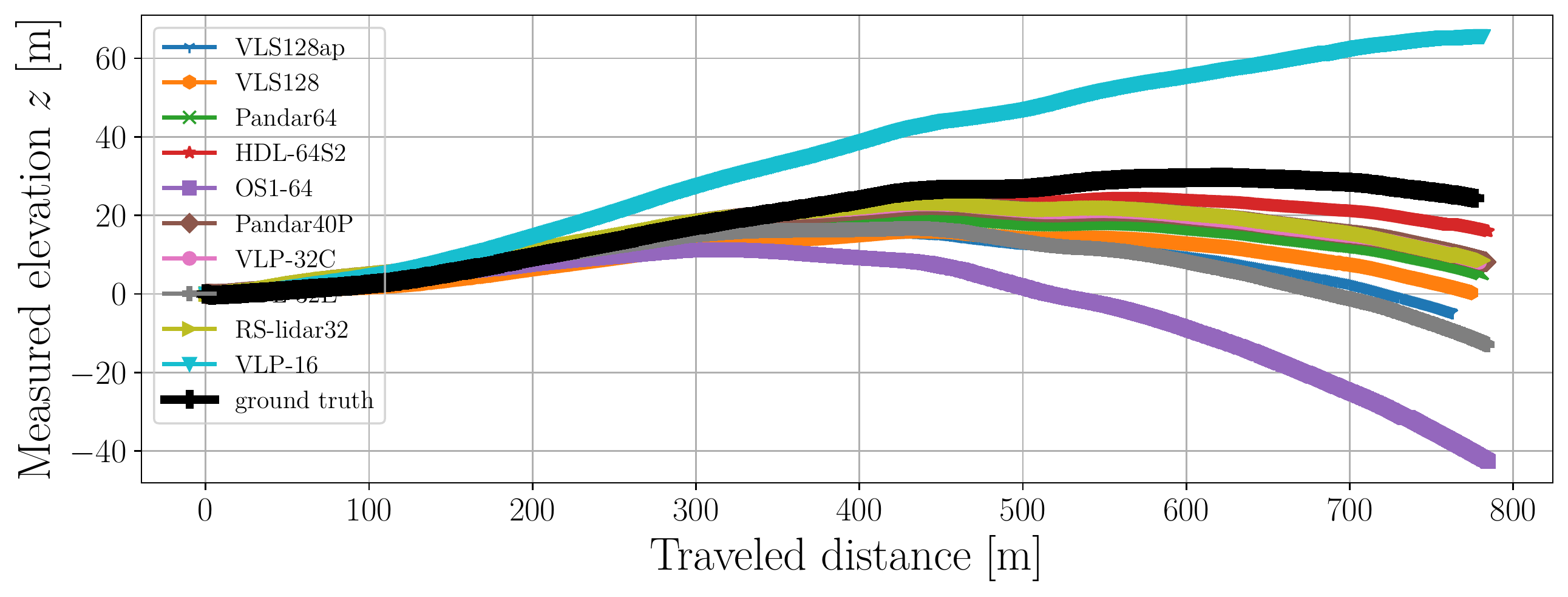}
		\label{F:elevation-c}
	}
	\caption[]{Elevation errors due error accumulation during map creation.}
	\label{F:elevation}
	\vspace{-1em}
\end{figure}
\section{Mapping Evaluation}
\label{s:mapping}
For each LiDAR and for each route, we created a 3D map using NDT for mapping implementation in Autoware. The NDT resolution was defined as 1\,m, the maximum number of iterations to 50, the minimum range was 3\,m and maximum range 200\,m. To add new scans, especially during congested traffic, the minimum shift (translation) parameter was set to 1\,m. After mapping the maps were not down-sampled to keep their integrity for this analysis, thus some of the maps are too large to be used for practical self-driving. The specs of the computer used in this evaluation are: Intel Xeon CPU E3-1545M with 4 cores/8 threads, NVIDIA Quadro M1000M GPU with 4GB GPU memory and 512 GPU cores, 32GB main RAM and a external 2TB SSD storage for log data recording.

Fig.~\ref{F:elevation} shows the elevation versus traveled distance for each LiDAR and each route while creating maps (ground truth is shown in black). Since each ride is slightly different (ex., changing lanes to avoid obstacles) the differences in trajectories in the XY plane are not considered here, instead the most important problem of drifting in elevation due to accumulated errors. While LiDARs with higher number of beams, finer vertical angular resolution, and higher range precision are closer to the ground truth, VLP-16 with lowest beam count and OS1-64 with lower range accuracy have the highest vertical errors. The reasons for this elevation error may be found on the NDT mapping algorithm itself, as reported previously in \cite{carballo2018end}, and more research is necessary, however this work sheds light on the dependency of the LiDAR beams configuration. The environment may also play a role in such errors, routes B are C are regular roads while route A is a narrow road on the university campus. 

We also consider as additional metrics to analyze map quality, the mean map entropy (MME) and the mean plane variance (MPV) discussed in Razlaw\etal\cite{razlaw2015evaluation}. The mean map entropy (MME) score $H(\bm{M})$ of map $\bm{M}$ is given by Eq.~\ref{eq:mme} as:
\begin{eqnarray}
	h(\bm{x}_k) &=& \frac{1}{2}\ln{|2\pi{e}\bm{\Sigma}(\bm{x}_k)|} \nonumber \\
	H(\bm{M}) &=& \frac{1}{M}\sum_{i}^{M}h(\bm{x}_i) \label{eq:mme}
\end{eqnarray}
The covariance $\bm{\Sigma}(\bm{x}_k)$ of point $\bm{x}_k$ is computed using a kd-tree neighborhood seearch with radius $r$. The mean plane variance (MPV) score $V(\bm{M})$ of the map is given by equation Eq.~\ref{eq:mpv} as:
\begin{eqnarray}
V(\bm{M}) &=& \frac{1}{M}\sum_{i}^{M}v(\bm{x}_i) \label{eq:mpv} 
\end{eqnarray}
where $v(\bm{x}_i)$ is the upper quartile of distances with the best fitting plane for the points around $\bm{x}_k$ in the search radius $r$.

For the ground truth reference map provided by MMS, the MME score is 0.294893 and the MPV score is 0.669889. Table~\ref{tab:mapping} summarizes the results during mapping\footnote{At the time of submission, missing MME and MPV values were being computed.}, including the number of points in each final map, the driving time for the route, number of scans included, and so on. As expected, LiDARs with higher number of beams take more time (iterations), while the fitness score remains similar for all sensors. The MME and MPV scores seem to favour the legacy sensors HDL-64S2 and HDL-32E, possibly due to the quality of their calibration and measurement accuracy when compared with newer devices. 

\begin{figure}[!htb]
	\centering
	\subfloat[][]{
		\includegraphics[width=0.3\textwidth]{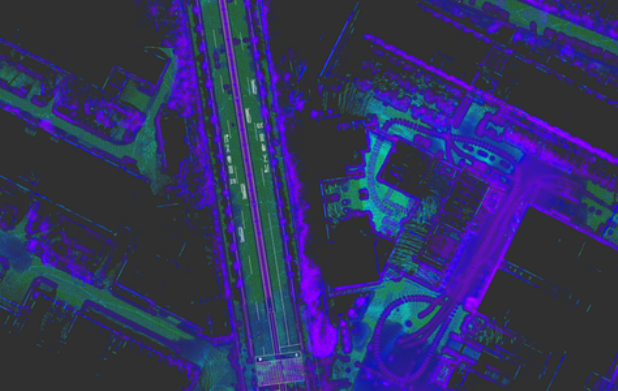}
		\label{F:mms-mme-a}
	}\\
	\subfloat[][]{
		\includegraphics[width=0.3\textwidth]{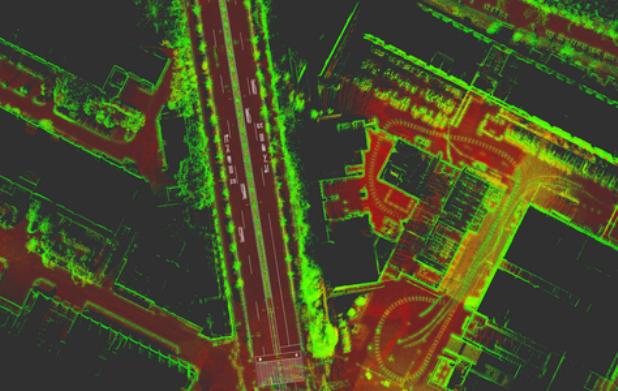}
		\label{F:mms-mme-b}
	}
	\caption[]{Reference MMS map coloured entropy\subref{F:mms-mme-a} (green lower entropy, magenta is higher entropy) and by plane variance \subref{F:mms-mme-b} (red lower variance, green higher variance) for each point.}
	\label{F:mms-mme}
	\vspace{-1em}
\end{figure}

One element which affects the MME and MPV scores is the presence of objects in the map, in particular on the road surface. Fig.~\ref{F:mms-mme} shows the reference MMS map colored by point entropy and point plane variance, both using a search radius $r$ of 1\,m. The map portion covered by vector maps was curated and no motion artifacts are observed, however map data inside the Nagoya University campus clearly show trails from objects in motion during mapping and not filtered here. The traffic conditions were slightly different while driving with each sensor, sometimes there were vehicles present leaving such trails on the map, and thus MME and MPV scores rose. Using the vector maps to filter road points and then removing points inside the road with high entropy and plane variance is a possible filtering strategy to consider.

\begin{figure*}[!htb]
	\begin{center}
		\footnotesize
		\setlength{\tabcolsep}{1pt}
		\begin{tabular}{m{0.1cm}c@{\hspace*{-5pt}}c@{\hspace{5pt}}c@{\hspace*{-5pt}}c@{\hspace{5pt}}c@{\hspace*{-5pt}}cm{0.1cm}}
			& \multicolumn{2}{c}{
				\hspace*{-5pt}\begin{minipage}{.3\linewidth}
					\centering
					\includegraphics[width=0.9\textwidth]{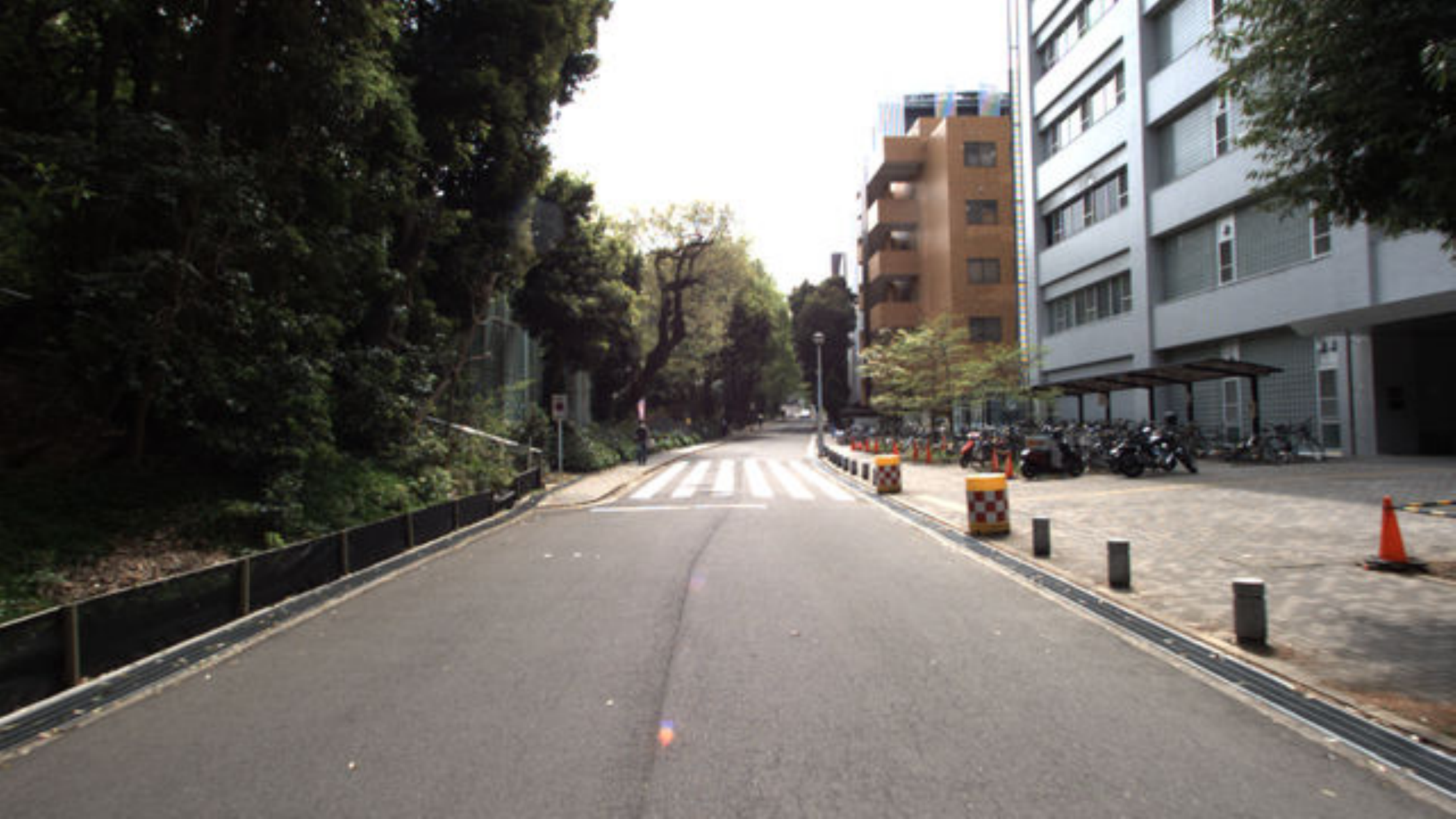}
				\end{minipage}
			} & \multicolumn{2}{c}{
				\hspace*{-5pt}\begin{minipage}{.3\linewidth}
					\centering
					\includegraphics[width=0.9\textwidth]{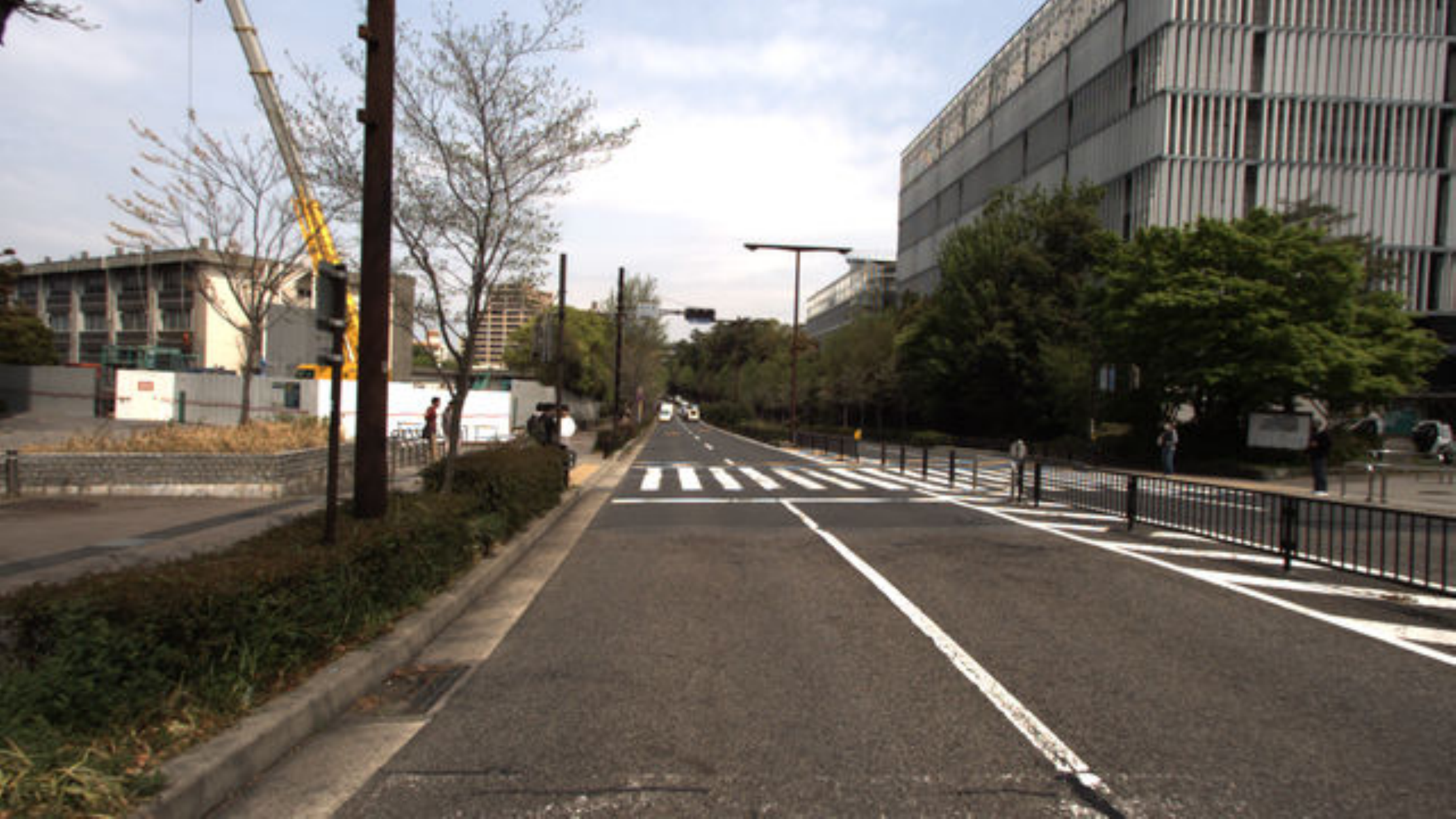}
				\end{minipage}
			} & \multicolumn{2}{c}{
				\hspace*{-5pt}\begin{minipage}{.3\linewidth}
					\centering
					\includegraphics[width=0.9\textwidth]{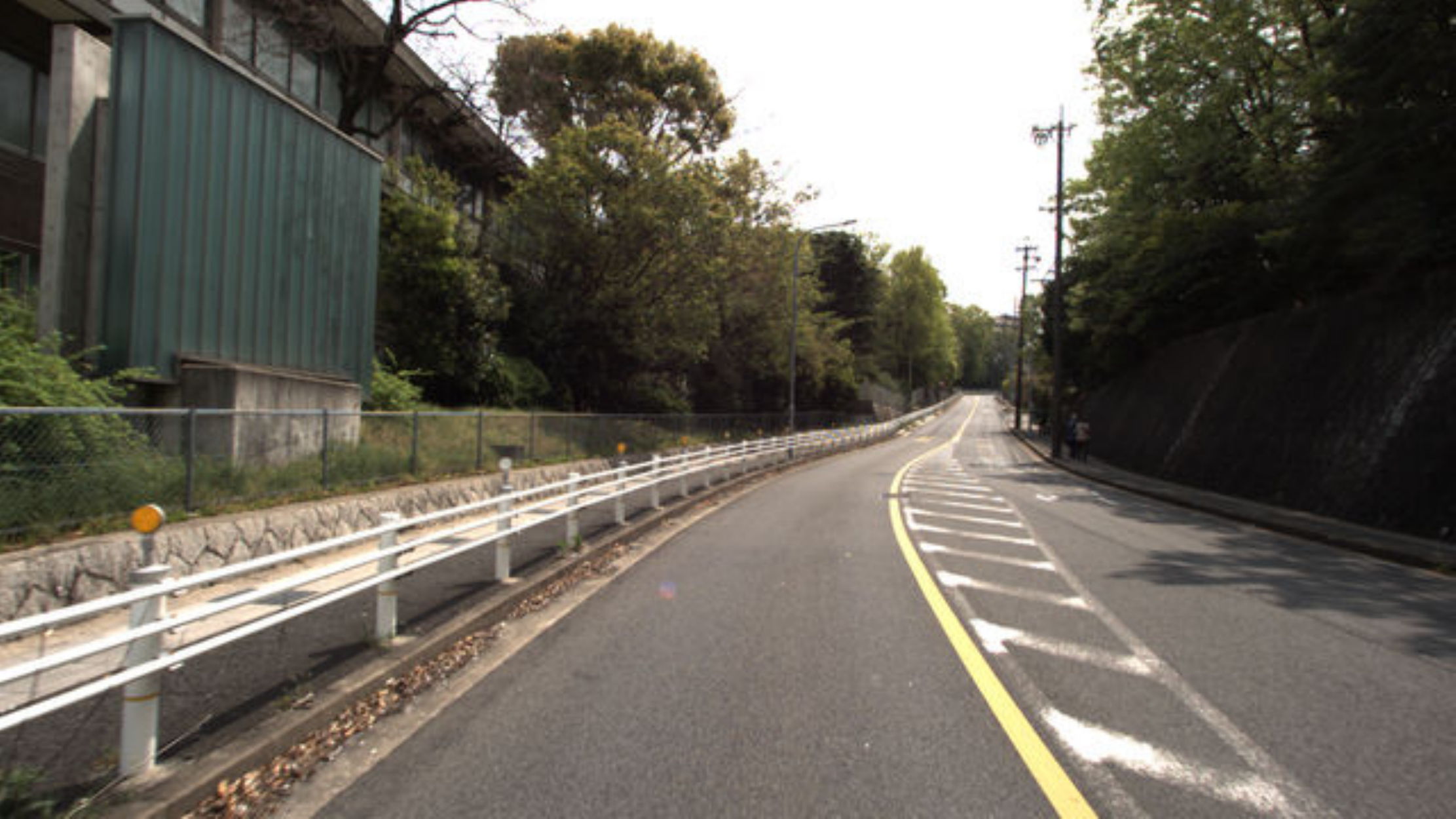}
				\end{minipage}
			} & \\
			\rotatebox{90}{\mbox{\normalsize{VLS-128AP}}} &
			\begin{minipage}{.15\linewidth}
				\centering
				\includegraphics[width=0.9\textwidth]{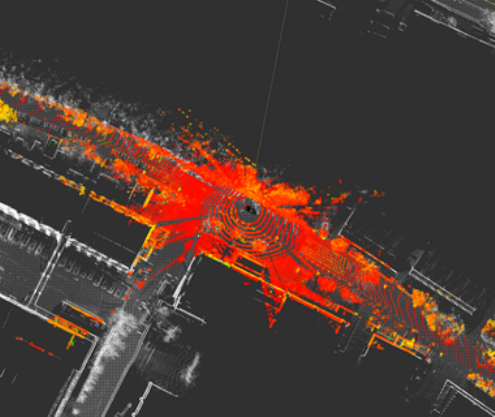}
			\end{minipage} & 
			\begin{minipage}{.15\linewidth}
				\centering
				\includegraphics[width=0.9\textwidth]{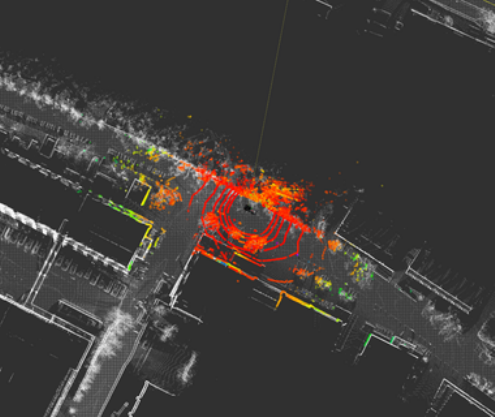}
			\end{minipage} & 
			\begin{minipage}{.15\linewidth}
				\centering
				\includegraphics[width=0.9\textwidth]{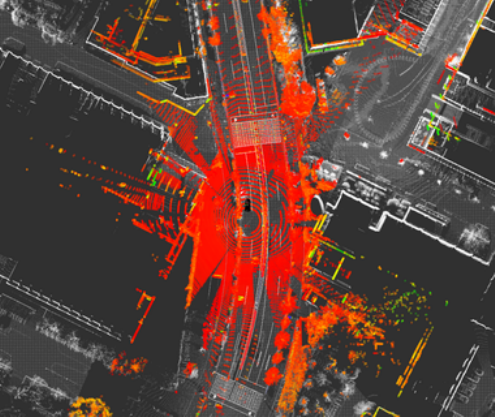}
			\end{minipage} &
			\begin{minipage}{.15\linewidth}
				\centering
				\includegraphics[width=0.9\textwidth]{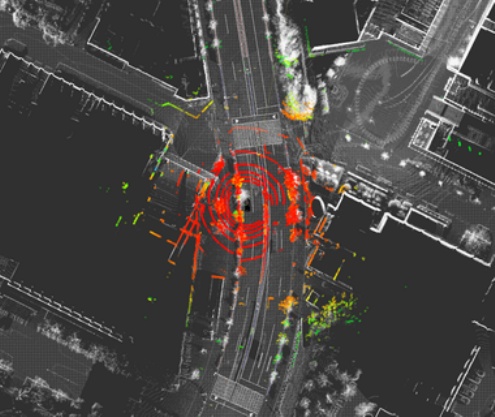}
			\end{minipage} &
			\begin{minipage}{.15\linewidth}
				\centering
				\includegraphics[width=0.9\textwidth]{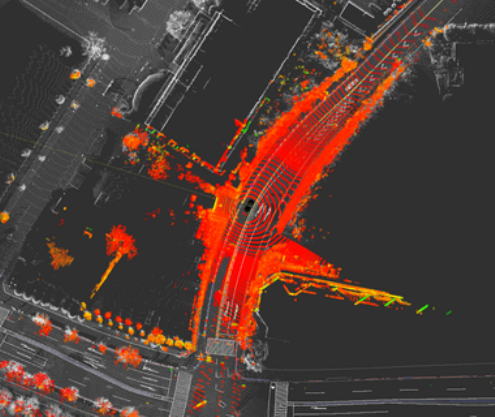}
			\end{minipage} &
			\begin{minipage}{.15\linewidth}
				\centering
				\includegraphics[width=0.9\textwidth]{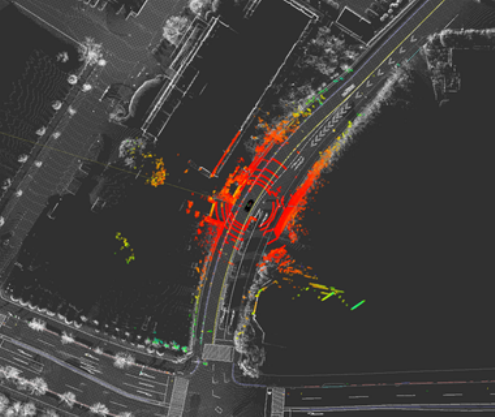}
			\end{minipage} & \rotatebox{90}{\mbox{\normalsize{VLP-16}}}\\
			\rotatebox{90}{\mbox{\normalsize{VLS-128}}} &
			\begin{minipage}{.15\linewidth}
				\centering
				\includegraphics[width=0.9\textwidth]{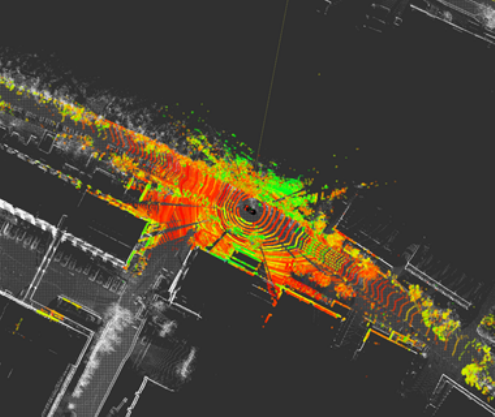}
			\end{minipage} & 
			\begin{minipage}{.15\linewidth}
				\centering
				\includegraphics[width=0.9\textwidth]{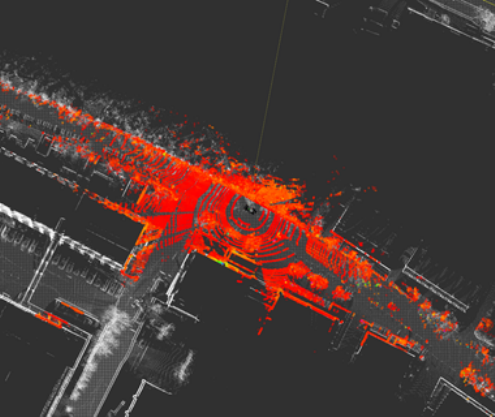}
			\end{minipage} & 
			\begin{minipage}{.15\linewidth}
				\centering
				\includegraphics[width=0.9\textwidth]{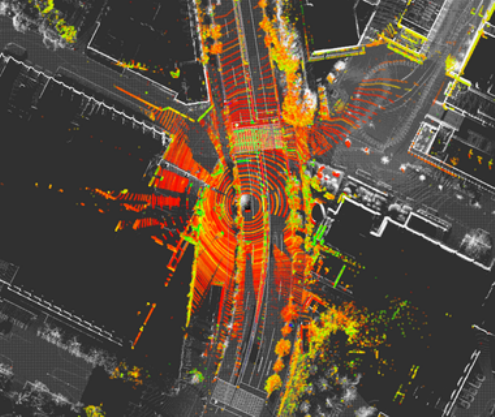}
			\end{minipage} &
			\begin{minipage}{.15\linewidth}
				\centering
				\includegraphics[width=0.9\textwidth]{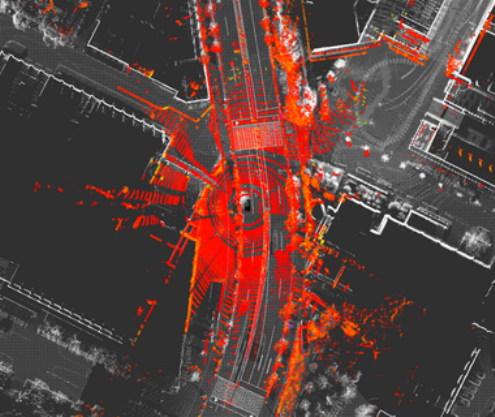}
			\end{minipage} &
			\begin{minipage}{.15\linewidth}
				\centering
				\includegraphics[width=0.9\textwidth]{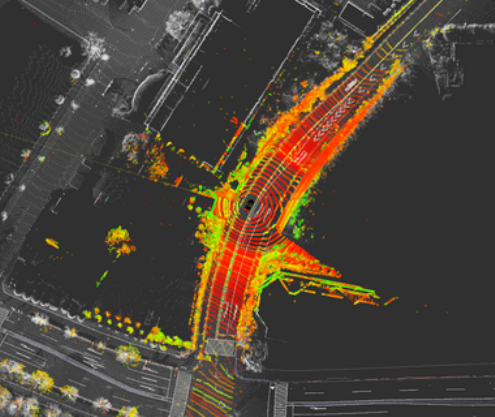}
			\end{minipage} &
			\begin{minipage}{.15\linewidth}
				\centering
				\includegraphics[width=0.9\textwidth]{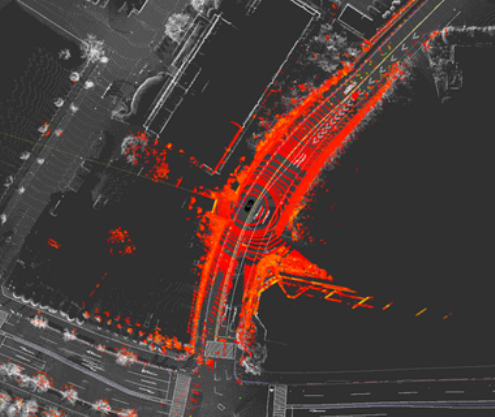}
			\end{minipage} & \rotatebox{90}{\mbox{\normalsize{Pandar-64}}}\\
			\rotatebox{90}{\mbox{\normalsize{HDL-64S2}}} &
			\begin{minipage}{.15\linewidth}
				\centering
				\includegraphics[width=0.9\textwidth]{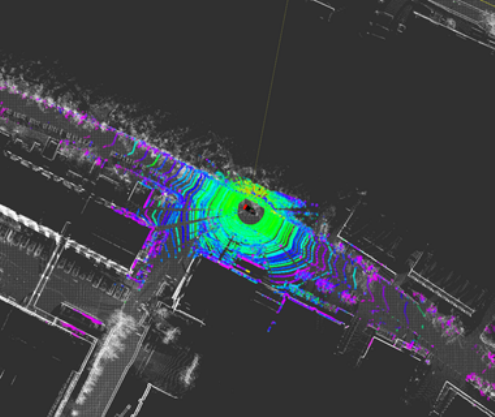}
			\end{minipage} & 
			\begin{minipage}{.15\linewidth}
				\centering
				\includegraphics[width=0.9\textwidth]{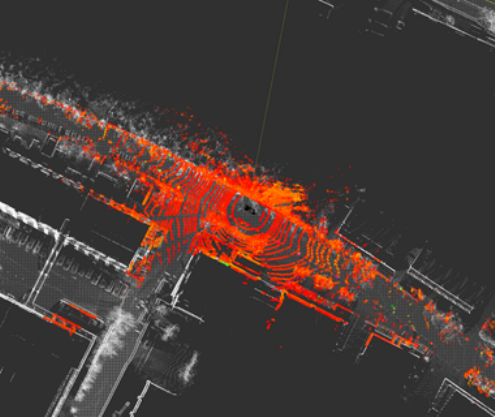}
			\end{minipage} & 
			\begin{minipage}{.15\linewidth}
				\centering
				\includegraphics[width=0.9\textwidth]{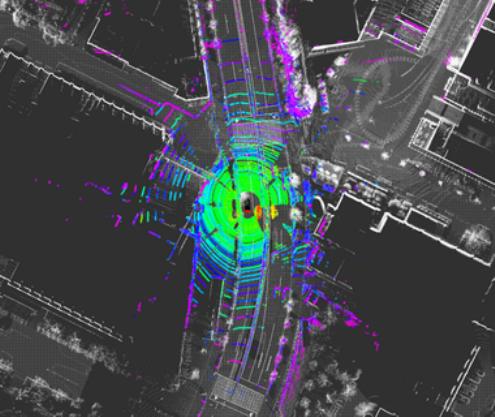}
			\end{minipage} &
			\begin{minipage}{.15\linewidth}
				\centering
				\includegraphics[width=0.9\textwidth]{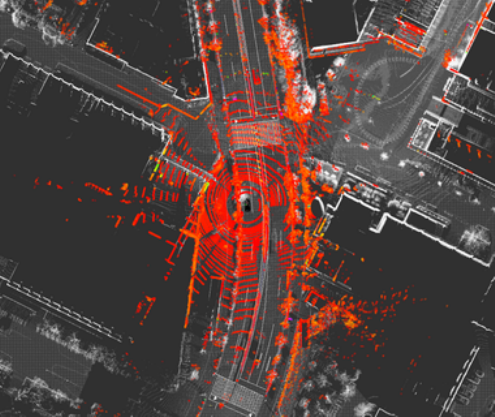}
			\end{minipage} &
			\begin{minipage}{.15\linewidth}
				\centering
				\includegraphics[width=0.9\textwidth]{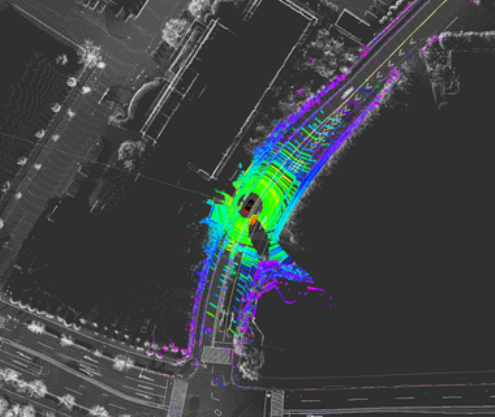}
			\end{minipage} &
			\begin{minipage}{.15\linewidth}
				\centering
				\includegraphics[width=0.9\textwidth]{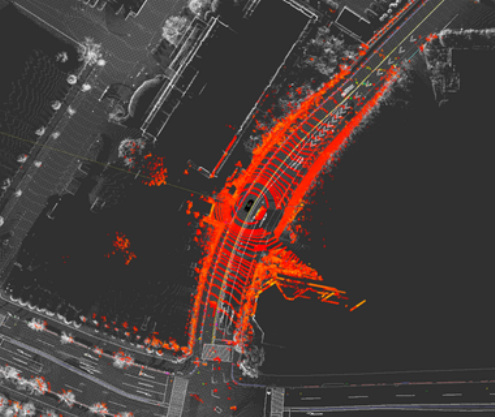}
			\end{minipage} & \rotatebox{90}{\mbox{\normalsize{Pandar-40p}}}\\
			\rotatebox{90}{\mbox{\normalsize{HDL-32E}}} &
			\begin{minipage}{.15\linewidth}
				\centering
				\includegraphics[width=0.9\textwidth]{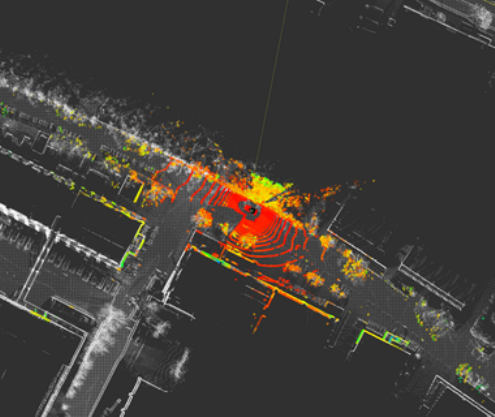}
			\end{minipage} & 
			\begin{minipage}{.15\linewidth}
				\centering
				\includegraphics[width=0.9\textwidth]{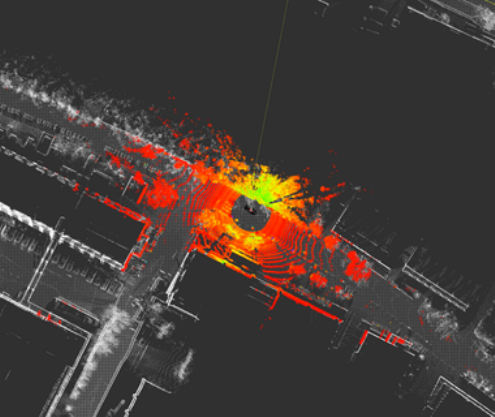}
			\end{minipage} & 
			\begin{minipage}{.15\linewidth}
				\centering
				\includegraphics[width=0.9\textwidth]{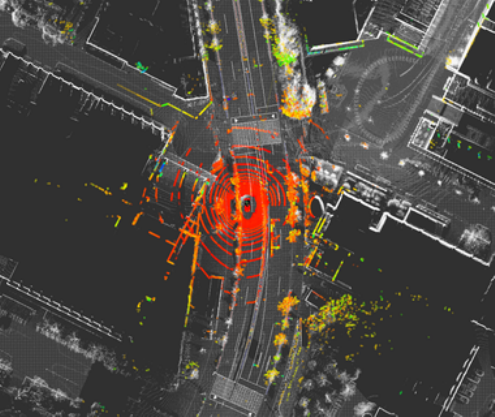}
			\end{minipage} &
			\begin{minipage}{.15\linewidth}
				\centering
				\includegraphics[width=0.9\textwidth]{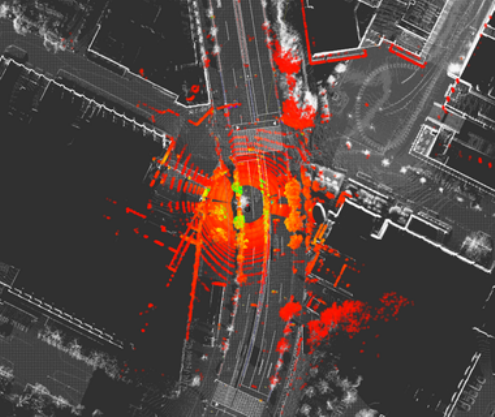}
			\end{minipage} &
			\begin{minipage}{.15\linewidth}
				\centering
				\includegraphics[width=0.9\textwidth]{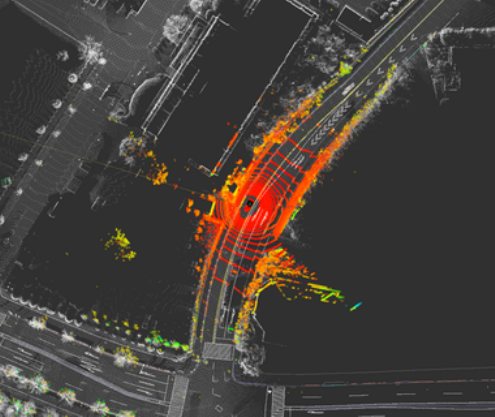}
			\end{minipage} &
			\begin{minipage}{.15\linewidth}
				\centering
				\includegraphics[width=0.9\textwidth]{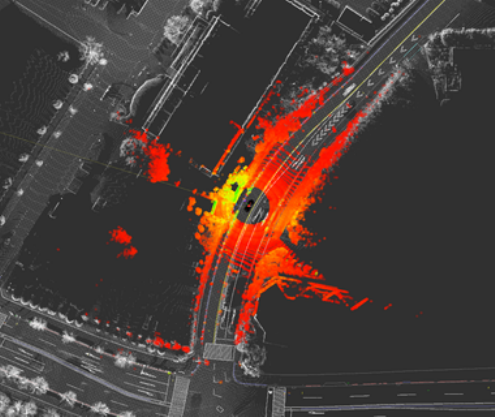}
			\end{minipage} & \rotatebox{90}{\mbox{\normalsize{OS1-64}}}\\
			\rotatebox{90}{\mbox{\normalsize{VLP-32C}}} &
			\begin{minipage}{.15\linewidth}
				\centering
				\includegraphics[width=0.9\textwidth]{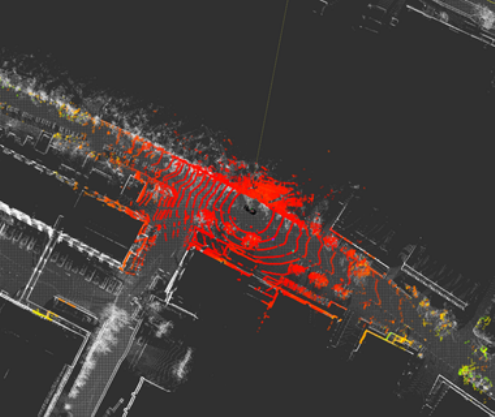}
			\end{minipage} & 
			\begin{minipage}{.15\linewidth}
				\centering
				\includegraphics[width=0.9\textwidth]{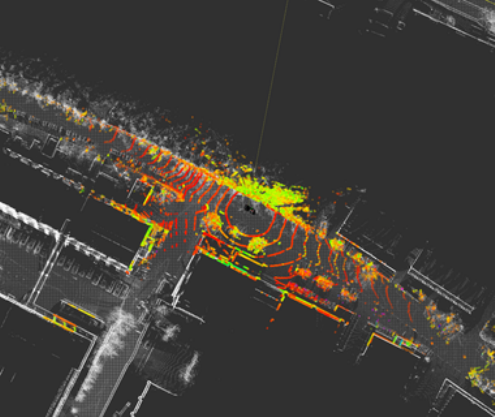}
			\end{minipage} & 
			\begin{minipage}{.15\linewidth}
				\centering
				\includegraphics[width=0.9\textwidth]{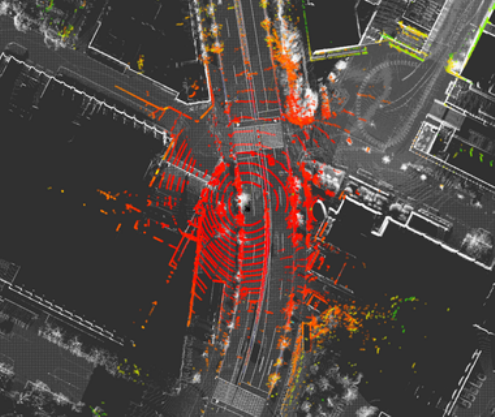}
			\end{minipage} &
			\begin{minipage}{.15\linewidth}
				\centering
				\includegraphics[width=0.9\textwidth]{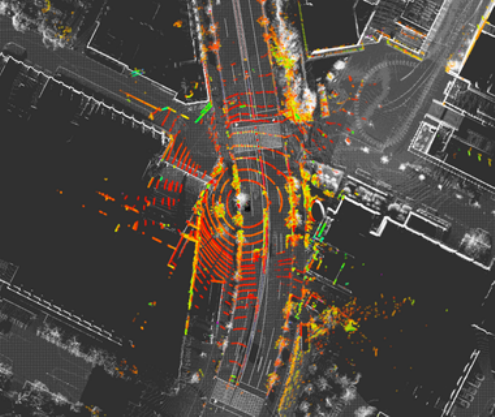}
			\end{minipage} &
			\begin{minipage}{.15\linewidth}
				\centering
				\includegraphics[width=0.9\textwidth]{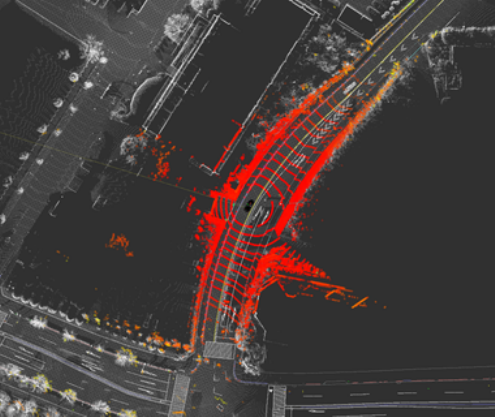}
			\end{minipage} &
			\begin{minipage}{.15\linewidth}
				\centering
				\includegraphics[width=0.9\textwidth]{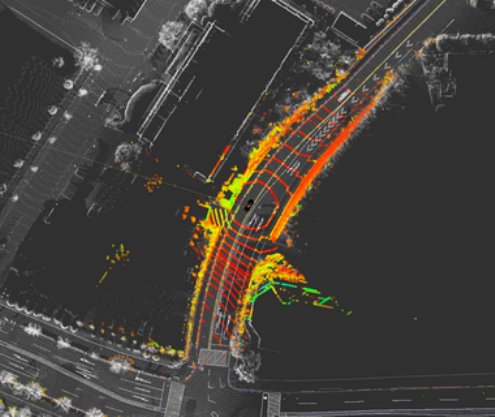}
			\end{minipage} & \rotatebox{90}{\mbox{\normalsize{RS-Lidar32}}}\\
			& \multicolumn{2}{c}{Route A} & \multicolumn{2}{c}{Route B} & \multicolumn{2}{c}{Route C} &  %& \multicolumn{2}{c}{LeiShen}\\
		\end{tabular}
	\end{center}
	\caption{Qualitative results of localization for each LiDAR for each LiDAR, grouped by route and colored by intensity (auto scaled), scenes of the chosen location are included. For each group, the LiDAR models are indicated by the given labels.}
	\label{F:loc-all}
	\vspace{-2em}
\end{figure*}

\section{Localization Evaluation}
\label{s:locoalization}

\begin{figure}[!htb]
	\centering
	\subfloat[][200\,m]{
		\includegraphics[width=0.18\textwidth]{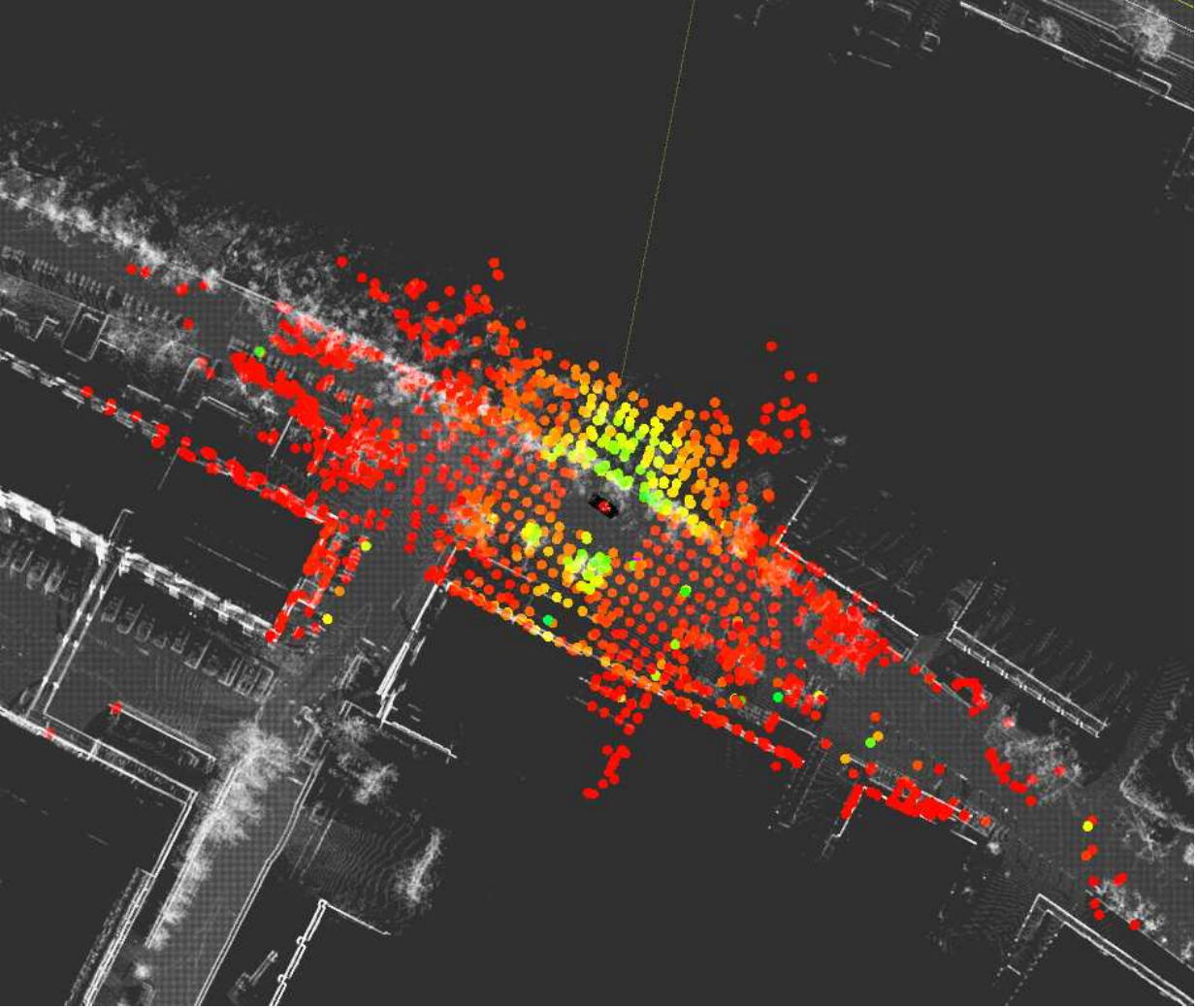}
		\label{F:voxelgrid-a}
	}
	\subfloat[][100\,m]{
		\includegraphics[width=0.18\textwidth]{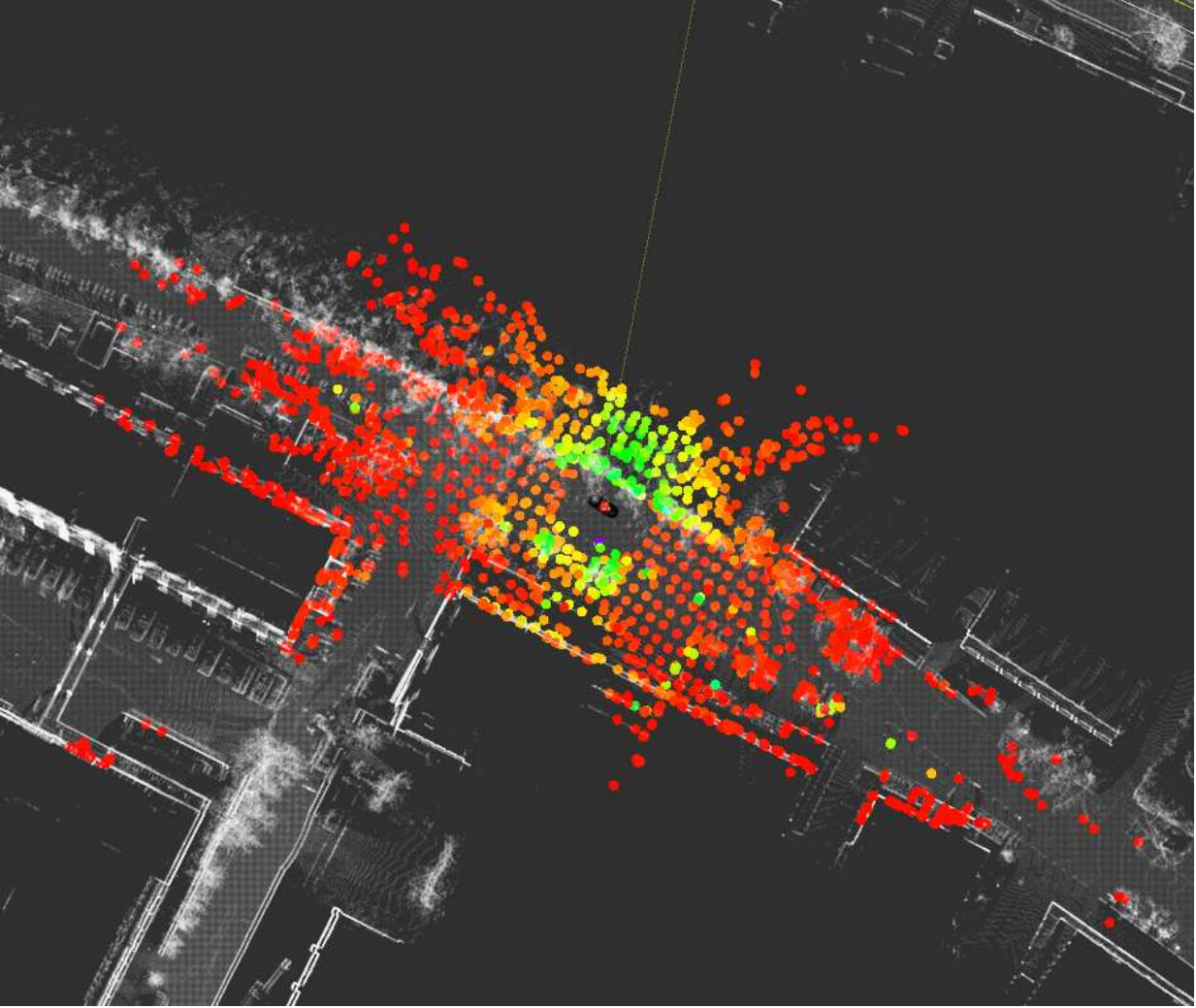}
		\label{F:voxelgrid-b}
	}\\
	\subfloat[][50\,m]{
		\includegraphics[width=0.18\textwidth]{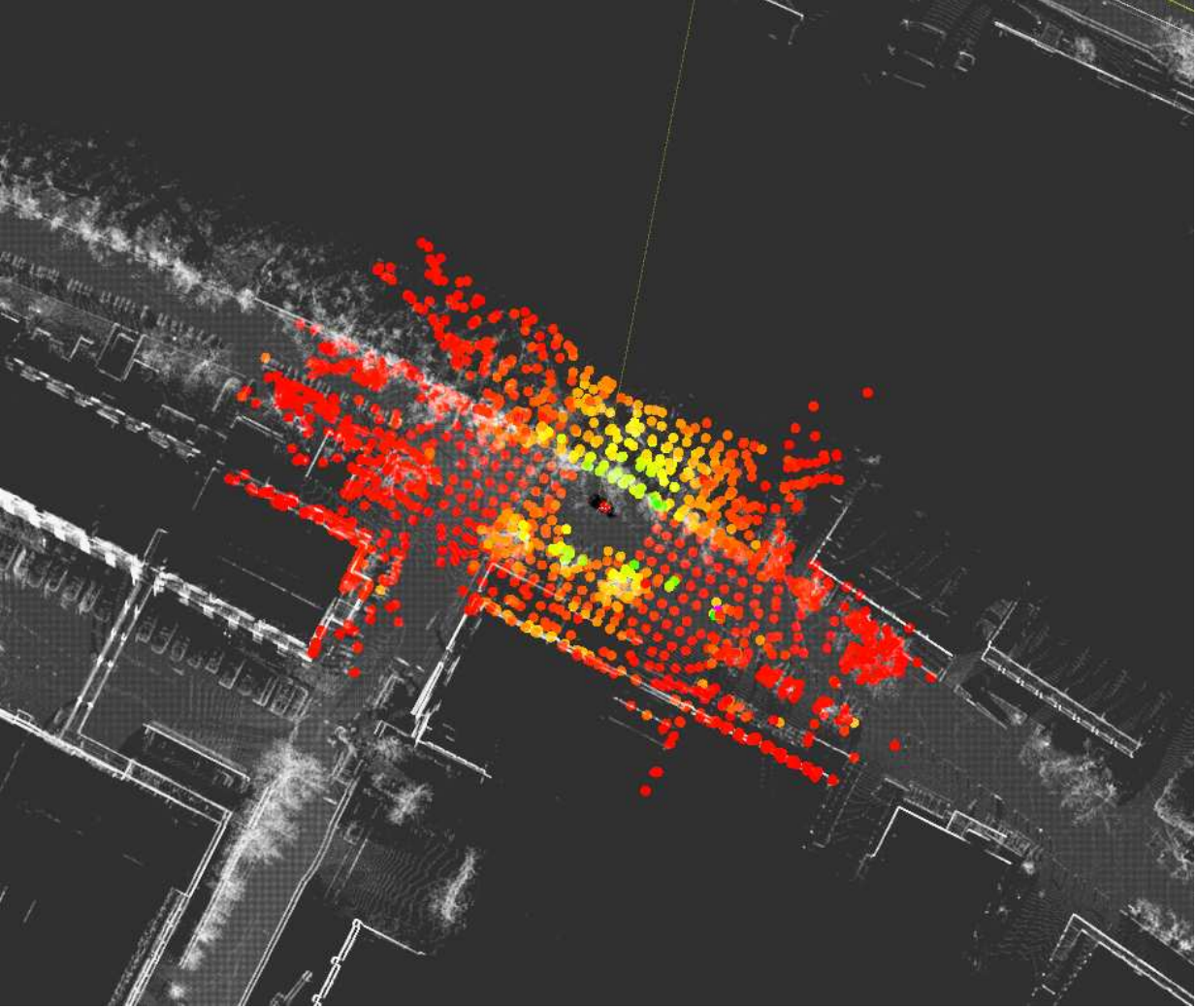}
		\label{F:voxelgrid-c}
	}
	\subfloat[][20\,m]{
		\includegraphics[width=0.18\textwidth]{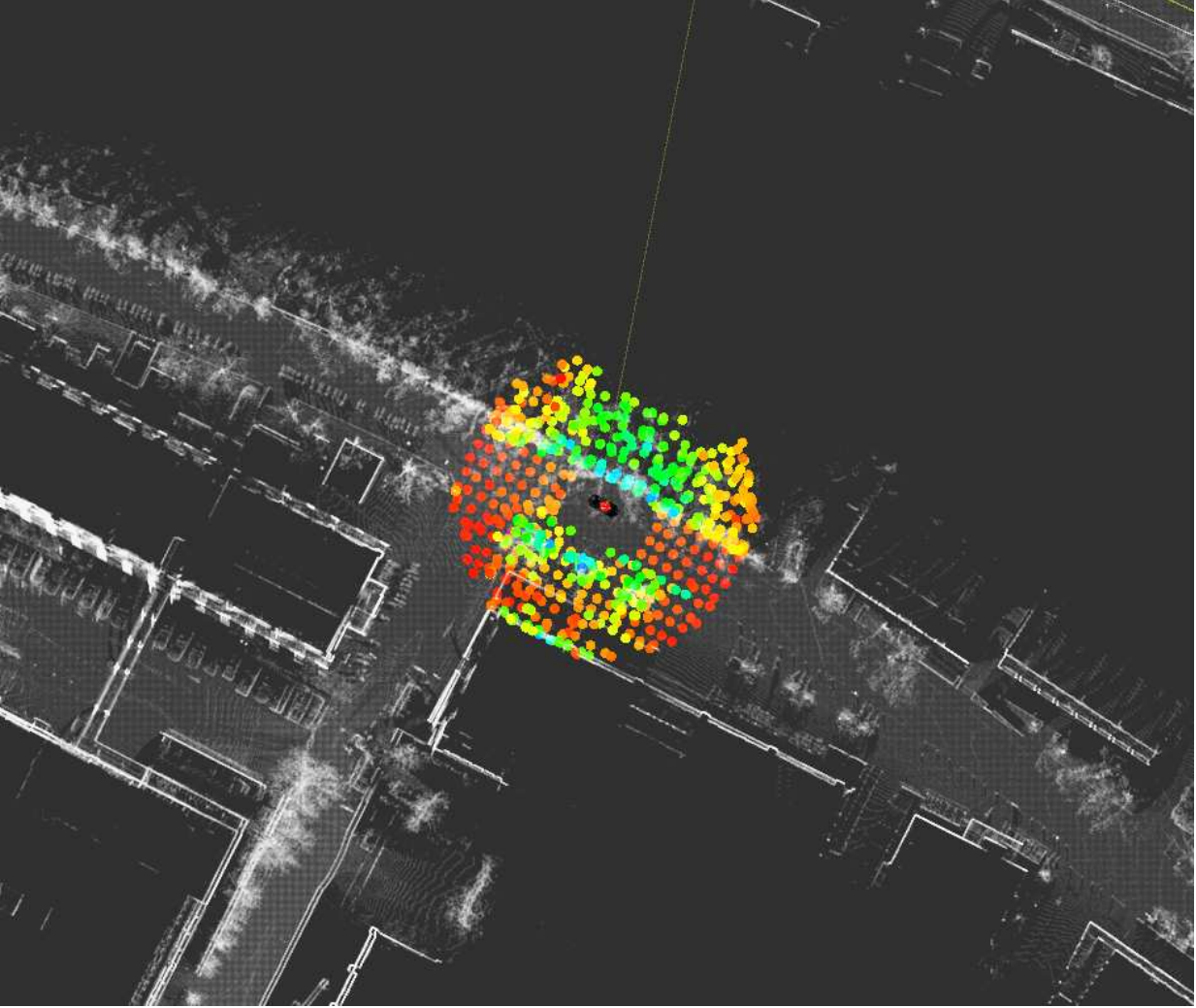}
		\label{F:voxelgrid-d}
	}
	\caption[]{Voxel grid filter applied to the OS1-64 pointcloud with different maximum range filters.}
	\label{F:voxelgrid}
	\vspace{-2em}
\end{figure}

We performed thirty experiments on localization, using the ground truth reference map, and NDT matching implementation in Autoware. NDT maching requires the input cloud to be down-sampled, we used the voxel grid filter with a voxel size of 2\,m and the maximum distance of 200\,m. Similar to NDT mapping, NDT resolution was set to 1\,m, maximum number of iterations to 50, error threshold was 1\,m. The initial position was manually defined for each case. Fig.~\ref{F:loc-all} show qualitative results of the different LiDARs during localization at specific locations. These images show the achieved coverage by each LiDARs's original input cloud after being transformed. Examples of filtered input cloud by the voxel grid filter are shown in Fig.~\ref{F:voxelgrid}.

\begin{figure}[!htb]
	\centering
	\subfloat[][Route A]{
		\includegraphics[width=0.35\textwidth]{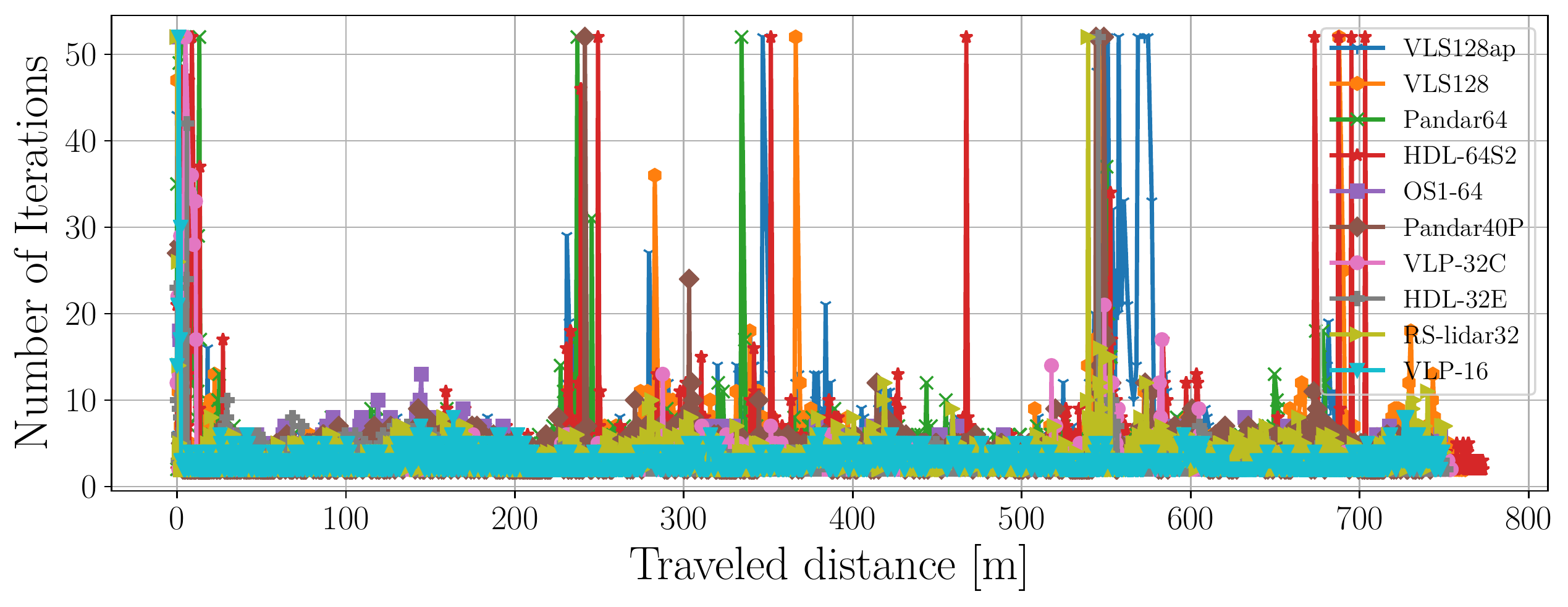}
		\label{F:localizationres-a}
	}\\
	\subfloat[][Route B]{
		\includegraphics[width=0.35\textwidth]{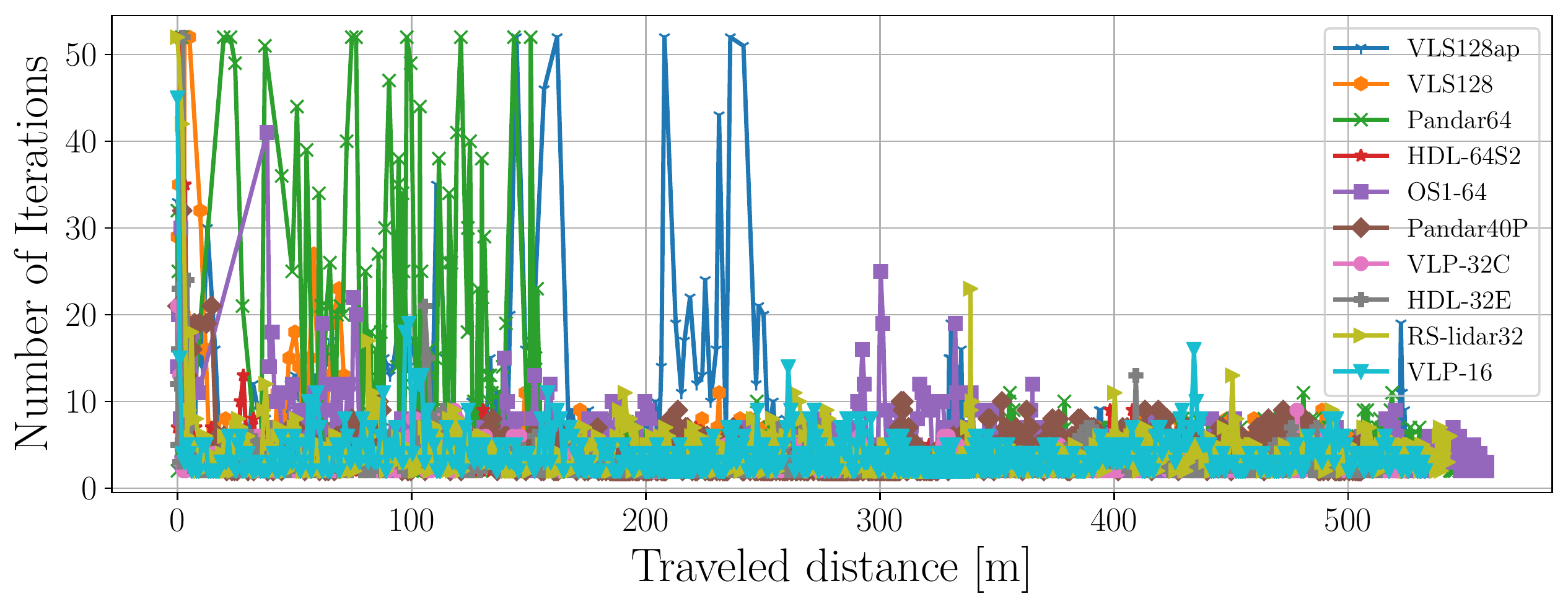}
		\label{F:localizationres-b}
	}\\
	\subfloat[][Route C]{
		\includegraphics[width=0.35\textwidth]{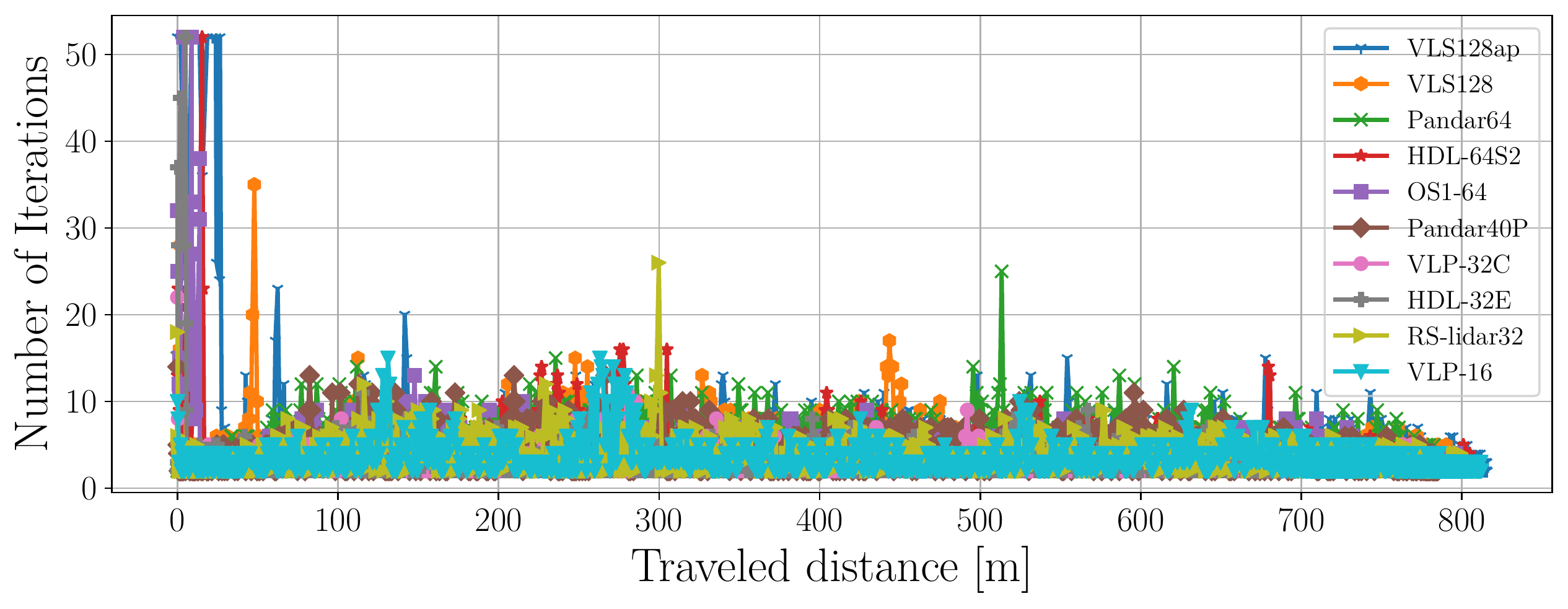}
		\label{F:localizationres-c}
	}
	\caption[]{Number of iterations per LiDAR and per route.}
	\label{F:localizationres}
	\vspace{-1em}
\end{figure}

Fig.~\ref{F:localizationres} shows the localization performance of the different LiDARs per each route. In general, VLP-16 had a very fast convergence while the LiDARs with higher number of beams struggled. The higher iterations at the beginning are due to the different initialization configurations required for each sensor and route. Route B shows the higher number of fluctuations in iterations, this is partially due to changes in the environment not included in the reference MMS map. Also, this route includes portions where maximum speed was 50\,km/h, the higher speed and the constant 10\,Hz scanning rate of the sensors points to the need of motion compensation in NDT. 

%\section{Range Ablation}
%\label{s:ablation}
%
%For each of the thirty localization experiments explained in Section~\ref{s:locoalization}, we tested 4 different configurations of the maximum distance of the voxel grid filter while keeping the voxel size as 2\.m. 
%voxel grid filter 200\,m, 100\,m, 50\,m and 20\,m

\section{Conclusions}
\label{s:concl}
In this work, we presented a detailed comparison and characteristics of ten different 3D LiDARs, each one a different model from diverse manufacturers, for navigation tasks (3D mapping and 6-DOF localization), using as common reference the Normal Distributions Transform (NDT). We analyzed each LiDAR pointcloud data alone and without assistance of any other navigation and dead reckoning systems. Data in this study comes from dynamic traffic data in our {\datasetname} dataset.

While ultimate reason may lie in the NDT implementation, the elevation during mapping for all 3D LiDARs was incorrect in all cases, drifting in elevation due to accumulated errors in posture estimation, the LiDAR's beams configuration and the surrounding environment seem to have an important effect. Correct localization was achieved for all LiDARs using a reference ground truth map. Even if all input clouds were equally filtered, the number of iterations varied largely especially for multiple beam configurations and also depending on the environment (existence of sufficient reference data). 

Future directions derived from this work include, studying the elevation problem in NDT mapping, evaluation using other algorithms such as LeGo-LOAM\cite{legoloam2018}, different down-sampling pre-processing filtering which preserves the point cloud structure, automatic tuning of localizer parameters. The application of MME based feature points for localization will be also considered.

\section*{Acknowledgments}
This work was supported by the Core Research for Evolutional Science and Technology (CREST) project of the Japan Science and Technology Agency (JST). 
We would like to extend our gratitude to the Autoware Foundation\footnote{\url{https://www.autoware.org}} to realize this project. Finally, and as a matter of course, this work would not have been possible without the invaluable support of Velodyne Lidar Inc., Ouster Inc., Hesai Photonics Technology Co., Ltd., and RoboSense--Suteng Innovation Technology Co., Ltd.

\bibliographystyle{IEEEtran}
\bibliography{IEEEabrv,multiple-lidars-and-ndt}

\end{document}